\documentclass{article}
\usepackage{genpa-preprint,times}

\usepackage{amsmath,amsfonts,bm}

\def\eqref#1{equation~\ref{#1}}

\def\1{\bm{1}}

\DeclareMathAlphabet{\mathsfit}{\encodingdefault}{\sfdefault}{m}{sl}
\SetMathAlphabet{\mathsfit}{bold}{\encodingdefault}{\sfdefault}{bx}{n}

\usepackage{hyperref}
\usepackage{url}
\usepackage{graphicx}
\usepackage{wrapfig}
\usepackage{needspace}
\usepackage{booktabs}
\usepackage{tabularx}
\usepackage{multirow}
\usepackage{pifont}
\usepackage[table]{xcolor}
\definecolor{GenPABlue}{HTML}{41769D}
\definecolor{GenPAOrange}{HTML}{D28B46}
\definecolor{GenPASection}{HTML}{EDF2F6}
\definecolor{TableVLM}{HTML}{F7EDDF}
\definecolor{TableVideo}{HTML}{EEEAF6}
\definecolor{TableImage}{HTML}{EAF0E2}
\newcommand{\robotpt}{\raisebox{0.1ex}{\scalebox{0.82}{\ding{51}}}}
\newcommand{\norobotpt}{\raisebox{0.1ex}{\scalebox{0.82}{\ding{55}}}}
\newcommand{\backbonegroup}[3]{\multicolumn{#1}{>{\columncolor{#2}}c}{\rule{0pt}{2.1ex}\textsc{#3}}\\}
\usepackage{flafter}
\usepackage{placeins}
\hypersetup{hidelinks,
  pdftitle={GenPA: Image Generators as Efficient Robot Policies},
  pdfauthor={Zian Deng, Tianyi Zhang, Zilong Zheng, Ziyu Liu, Tao Zhang, Yao Lyu, Shengbo Eben Li}}

\title{GenPA: Image Generators as\\Efficient Robot Policies}

\newcommand{\projecturl}{https://deng-zian.github.io/GenPA-website/}
\newcommand{\affilmark}[1]{\textsuperscript{#1}}
\author{%
  {\normalsize\bfseries
    Zian Deng\affilmark{1,2}\quad Tianyi Zhang\affilmark{3}\quad
    Zilong Zheng\affilmark{4}\quad Ziyu Liu\affilmark{5}\\[2pt]
    Tao Zhang\affilmark{5}\quad Yao Lyu\affilmark{3,\textdagger}\quad
    Shengbo Eben Li\affilmark{3,6,\textdagger}}\\[4pt]
  {\small
    \affilmark{1}College of Computer Science and Artificial Intelligence, Fudan University\\
    \affilmark{2}Beijing Zhongguancun Academy\quad
    \affilmark{3}School of Vehicle and Mobility, Tsinghua University\\
    \affilmark{4}Faculty of Mechanical Engineering, Delft University of Technology\quad
    \affilmark{5}SunRising AI Ltd.\\
    \affilmark{6}College of Artificial Intelligence, Tsinghua University\\
    \affilmark{\textdagger}Corresponding authors}
  \\[3pt]{\small\textbf{Project page}\enspace\expandafter\url\expandafter{\projecturl}}
}
\newcommand{\GenPALibero}{99.10}
\newcommand{\GenPACleanPooled}{83.79}
\newcommand{\GenPARoboTwinClean}{92.08}
\newcommand{\GenPARoboTwinRandom}{90.82}
\newcommand{\GenPARoboTwinAvg}{91.45}

\begin{document}
\maketitle
\begin{abstract}
Pretrained visual generation models have shown great promise in robotic
manipulation. However, a representative line of work still treats future
visual prediction as the central link between visual generative pretraining and
action learning, retaining it as a training objective even when future scenes
are no longer generated at inference time. This paradigm organizes policy
learning around the proxy task of predicting future visual states. We argue
that the instruction-following, spatial understanding, and visual imagination
capabilities acquired through image generation pretraining can support
control without future visual prediction: image generation models themselves
can be directly adapted into powerful action learners. {Building on this perspective, we propose
GenPA, which fine-tunes an image generation backbone with a lightweight
19M-parameter action regression adapter and reuses its native generative
objective for perception co-training. The native image-editing interface accommodates
different visual targets without task-specific perception heads, enabling
comparisons of current-scene perception, future prediction, and their
combinations within a common policy architecture.}
At deployment, GenPA requires only
a single backbone forward pass and action regression, without visual
generation or iterative action denoising. Without additional large-scale
robot-data pretraining, GenPA demonstrates strong control performance and
robustness to perturbations across simulation benchmarks
(\GenPALibero\% success on LIBERO and \GenPACleanPooled\% on LIBERO-Plus)
and real-world manipulation tasks.
Under compiled inference, GenPA takes 20.9\,ms per action chunk on a single
H100 GPU, achieving $1.7$--$3.4\times$ speedups over strong VLA and WAM baselines.
Our latency analysis further highlights architectural advantages of
visual generation backbones for efficient robot control.
Ablations across visual tasks
and their combinations identify current-depth supervision as a simple,
effective training recipe. Together, these results show that image generators
can be directly adapted into strong, efficient robot policies without requiring
future visual prediction.
\end{abstract}

\begin{figure}[t]
\centering
\includegraphics[width=0.98\linewidth]{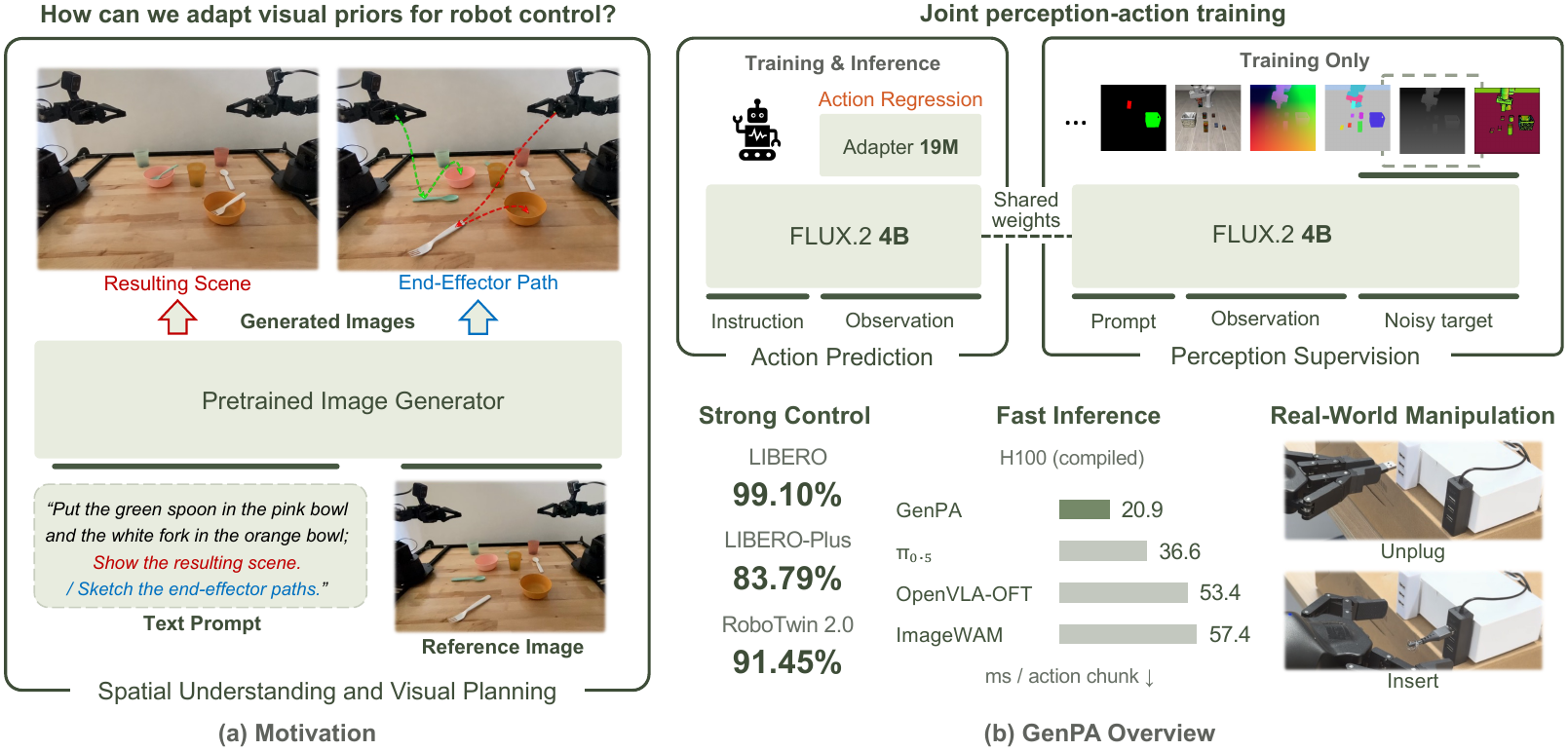}
\caption{\textbf{Motivation and overview of GenPA.}
\textbf{(a)} Given a scene image, a manipulation instruction,
and an editing request, GPT Image 2 visualizes a goal state or sketches
end-effector paths in image space, illustrating the potential of image
generation models for spatial understanding and visual planning
(generation details in Appendix~\ref{app:figure1-generation}).
\textbf{(b)} GenPA jointly trains action prediction and perception with
shared backbone weights and uses only the action path at deployment.
The action regression adapter has 19M parameters in the LIBERO configuration.}
\label{fig:genpa-overview}
\end{figure}

\section{Introduction}
\label{sec:introduction}

Recent advances in image generation pretraining have yielded strong
instruction-following and visual understanding capabilities. Beyond
synthesizing realistic images, these models can identify objects referred to
in language, follow spatial constraints, and visualize instructed changes to
a scene~\citep{xiao2025omnigen,blackforestlabs2025fluxkontext,zhao2026qwenimage2}.
As illustrated in Figure~\ref{fig:genpa-overview}(a), given a robot manipulation
instruction and a corresponding editing request, an image generation system
can depict the scene after task completion or sketch end-effector motion
paths on the current image. These behaviors suggest that the spatial
understanding and visual imagination capabilities acquired through
generative pretraining may provide valuable priors for robot action learning.

Pretrained image and video generation models have both been adapted for robot
control. A representative line of work treats future visual prediction as
the central link between generative pretraining and action learning,
using generated visual subgoals to guide goal-conditioned
policies~\citep{black2023susie} or decoding actions from predicted visual
transitions through inverse dynamics~\citep{li2026lingbotva}.
Several methods jointly predict future observations and actions within a
shared generative
model~\citep{shen2025videovla,kim2026cosmospolicy,ye2026dreamzero}.
Recent approaches further reduce or bypass future visual generation at
deployment while retaining future prediction as a co-training
objective~\citep{yuan2026fastwam,zhang2026imagewam,ye2026gigaworldpolicy}.
In the broader context of predictive policy learning, future supervision has
also been extended to depth, 3D pointmaps, and semantic
features~\citep{zhang2025dreamvla,yan2026flexpi}.
Despite differences in architecture and deployment, these methods continue
to organize policy learning around the proxy task of predicting the future.

We argue that the control value of image generation pretraining need not be
realized through the intermediate objective of future visual prediction: image
generation models themselves can be directly adapted into powerful action
learners. Prior work demonstrates the value of features from frozen image
generators for control~\citep{shin2026orca}. {We instead adapt the backbone
itself through action learning, with a lightweight adapter mapping its features
to actions.} Meanwhile, direct action learning need not discard
the generator's visual output capabilities. Research on generative perception
shows that depth estimation, surface normal prediction, and
segmentation can be learned as image generation tasks~\citep{gabeur2026visionbanana,han2026sensenovavision}.
Together with evidence that auxiliary depth and spatial
supervision benefit policy learning~\citep{li2025qdepth,tu2026sgvla}, these
{findings motivate reusing the generator's native interface for
current-scene perception co-training. Representing different targets through
this common interface also enables a systematic investigation of which visual
supervision benefits control, including comparisons with future prediction
and combinations of visual tasks.}

Building on this perspective, we introduce \textbf{GenPA} (\textbf{Gen}erative
\textbf{P}erception for \textbf{A}ction), an end-to-end policy that fine-tunes
the pretrained image generation backbone FLUX.2 klein Base
4B~\citep{blackforestlabs2026flux2klein} with a 19M-parameter action
regression adapter by default. During training, action and perception objectives jointly
update the backbone, with perception supervision reusing its native
image-editing interface and generative objective. Deployment requires only a single
backbone forward pass and action regression, without visual generation or
iterative action denoising. Figures~\ref{fig:genpa-overview}(b)
and~\ref{fig:genpa-architecture} provide an overview and the detailed
architecture, respectively.

Without additional large-scale robot-data pretraining, GenPA
achieves strong control performance on LIBERO, LIBERO-Plus, and
RoboTwin~2.0. On real-world contact-rich manipulation, GenPA outperforms
$\pi_{0.5}$, Fast-WAM, and ImageWAM while demonstrating strong robustness
to visual perturbations. Under compiled inference on a single H100 GPU,
GenPA takes 20.9\,ms per action chunk, achieving $1.7$--$3.4\times$ speedups
over the evaluated VLA and WAM baselines.
{Our ablations further show that direct action regression achieves
better closed-loop control performance with lower inference latency than a
flow-based action head in our main evaluations. Compared with direct depth
regression, native flow-based depth supervision improves control robustness
at a higher training cost.}

Our contributions are twofold. \textbf{(i)} We introduce GenPA,
demonstrating that an image generation backbone can be directly adapted into
a strong, efficient robot policy without a future visual prediction
objective. {Its native visual co-training interface supports diverse
supervision targets without task-specific perception heads, enabling systematic
comparisons of visual tasks and their combinations within a common policy
architecture.} \textbf{(ii)} {Our controlled studies show that strong
closed-loop control can be achieved with direct action regression, without
iterative action denoising, and identify current-depth supervision as a simple,
effective training recipe.} Our latency analysis further reveals
an architectural advantage of visual generation backbones for robot
control: lightweight observation encoding and independent language encoding
align computation with the different update frequencies of observations
and instructions.

\section{GenPA}
\label{sec:method}

\begin{figure}[t]
\centering
\includegraphics[width=\linewidth]{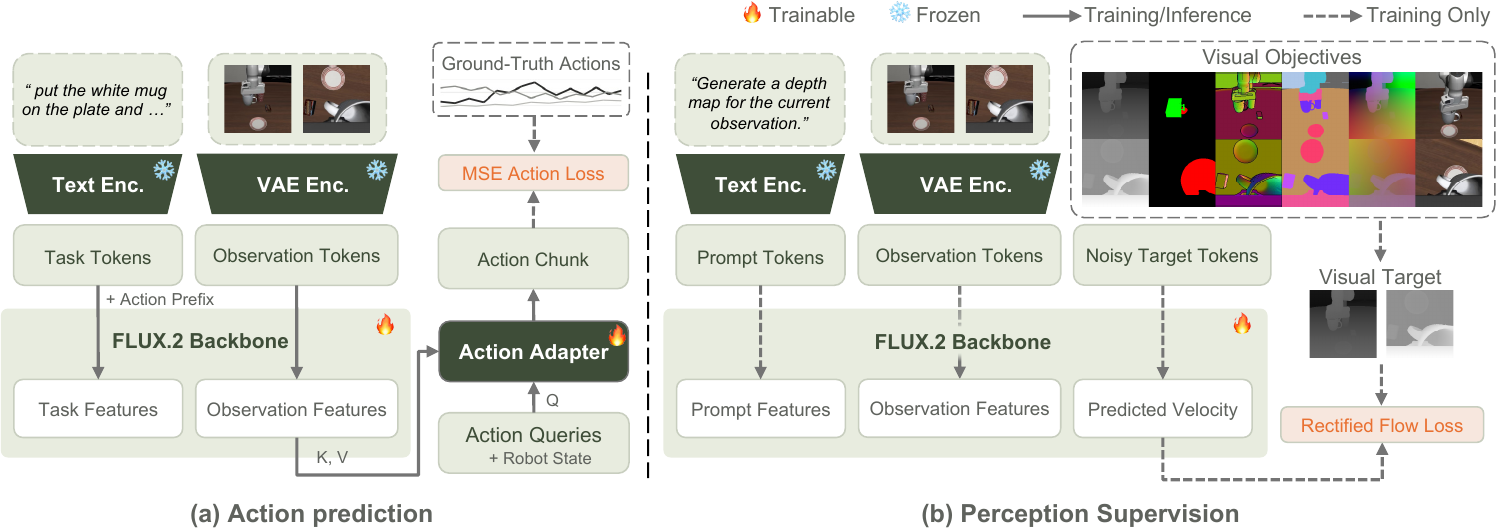}
\caption{\textbf{Architecture of GenPA.} \textbf{(a)} The action path extracts instruction-conditioned observation features from the image generation backbone. A lightweight regression adapter reads these features and predicts an action chunk conditioned on proprioception. \textbf{(b)} The perception path reuses the generator's native image-editing interface to learn image-form perceptual targets with a rectified-flow objective. The two paths use separate forward passes with shared backbone weights. Only the action path is used at inference.}
\label{fig:genpa-architecture}
\end{figure}

At robot time step $t$, given current RGB observations $\mathbf{o}_t$,
a language instruction $\ell$, and proprioception $\mathbf{s}_t$,
GenPA predicts an action chunk
$\mathbf{A}_t=(\mathbf{a}_t,\ldots,\mathbf{a}_{t+H-1})$ of length $H$.
As shown in Figure~\ref{fig:genpa-architecture}, GenPA predicts actions from
the image generation backbone's final reference-token features and
reuses its native image-editing interface for perception supervision.
Action and perception training use separate forward passes with shared
backbone parameters.

\subsection{Preliminaries: Image-Conditioned Generation}
\label{sec:generative-interface}

We instantiate GenPA with FLUX.2 klein Base 4B, a pretrained rectified-flow
transformer supporting image generation and
editing~\citep{blackforestlabs2026flux2klein}. Its editing interface processes
three groups of tokens: text conditions, clean reference-image latents, and
noised target-image latents. The reference image supplies visual context;
the target represents the image to be generated. Let $E_{\mathrm{vae}}$
denote the VAE encoder. For a target image $\mathbf{y}$ with latent
$\mathbf{z}=E_{\mathrm{vae}}(\mathbf{y})$ and Gaussian noise
$\boldsymbol{\epsilon}\sim\mathcal{N}(0,I)$, the linear flow path
is~\citep{liu2023flowstraightfast}
\begin{equation}
\mathbf{z}_{\tau}=(1-\tau)\mathbf{z}+\tau\boldsymbol{\epsilon},
\qquad \tau\in[0,1].
\label{eq:visual-flow}
\end{equation}
Here $\tau$ is the flow time: $\tau=0$ denotes clean data, and $\tau=1$
denotes noise. The generator predicts the velocity
$\boldsymbol{\epsilon}-\mathbf{z}$ along this path; image synthesis requires
iterative integration followed by VAE decoding.

\subsection{Direct Action Adaptation}
\label{sec:action-path}

\paragraph{Adapting the backbone.}
We encode the current camera observations as reference latents
$\mathbf{z}_t^R=E_{\mathrm{vae}}(\mathbf{o}_t)$ and the manipulation
instruction as text embeddings $E_{\mathrm{text}}(\ell)$, where
$E_{\mathrm{text}}$ is the text encoder. A learned
Action Prefix $\mathbf{P}^{A}$ is prepended to the text embeddings to
provide action-specific conditioning. The action forward pass contains
only text and reference-image tokens, with the flow-time condition fixed
at $\tau_A=0$.
GenPA uses bidirectional attention among valid text tokens, including the
prefix, and reference tokens, allowing the reference features to incorporate
the manipulation instruction. Let $\theta$ denote the trainable generator
parameters, and $F_\theta^R$ the backbone computation that returns the final
reference-token features. The action path is
\begin{equation}
\begin{aligned}
\mathbf{h}_t^R
&=F_\theta^R\!\left([\mathbf{P}^{A};E_{\mathrm{text}}(\ell)],
                    \mathbf{z}_t^R;\tau_A=0\right),\\
\hat{\mathbf{A}}_t&=g_\psi(\mathbf{h}_t^R,\mathbf{s}_t).
\end{aligned}
\label{eq:action-path}
\end{equation}

\paragraph{Lightweight action readout.}
The action adapter $g_\psi$, with trainable parameters $\psi$, is a
four-layer Transformer decoder that uses
one learned query per action step to read the final reference features
through cross-attention. The queries are conditioned on proprioception,
and a linear head predicts the action chunk in parallel. The default adapter
contains 19M parameters, counted for the LIBERO configuration.
Over a demonstration dataset $\mathcal{D}$,
we optimize a mean-squared action loss
\begin{equation}
\mathcal{L}_{\mathrm{act}}
=\mathbb{E}_{\mathcal{D}}\!\left[
\left\|\hat{\mathbf{A}}_t-\mathbf{A}_t\right\|_{\mathrm{valid}}^2
\right],
\label{eq:action-objective}
\end{equation}
where the error is averaged over valid action elements in each demonstration
sample.

\subsection{Native Perception Supervision}
\label{sec:visual-supervision}

For perception co-training, we reuse FLUX's image-conditioned flow
formulation with a modified attention pattern.
For each perception task $k$, we pair the current observation with a task prompt
$\ell_k^V$ and an image-form perceptual target $\mathbf{y}_t^{(k)}$,
such as a depth map, surface normals, or segmentation. These targets are
encoded by the same VAE and represented as target-image tokens, without
introducing task-specific perceptual heads.

We encode each target as
$\mathbf{z}^{(k)}=E_{\mathrm{vae}}(\mathbf{y}_t^{(k)})$ and construct
$\mathbf{z}^{(k)}_\tau=(1-\tau)\mathbf{z}^{(k)}+\tau\boldsymbol{\epsilon}$
following Equation~\ref{eq:visual-flow}. The generator predicts the target
velocity conditioned on the reference observation and perception-task prompt.
In this forward pass, text and reference tokens attend to each other but
cannot attend to the noised target; target tokens can attend to all three
groups. Text and target tokens receive the sampled time condition $\tau$,
while the reference tokens receive a zero time condition.
Let $v_\theta$ denote the generator's velocity prediction for
target tokens. The perception objective is
\begin{equation}
\mathcal{L}_{\mathrm{vis}}^{(k)}
=\mathbb{E}_{\mathcal{D}_k,\tau\sim p(\tau),\boldsymbol{\epsilon}}\!\left[
w(\tau)\left\|
v_\theta\!\left(\mathbf{z}^{(k)}_\tau;
\mathbf{z}_t^R,E_{\mathrm{text}}(\ell_k^V),\tau\right)
-(\boldsymbol{\epsilon}-\mathbf{z}^{(k)})
\right\|_{\mathrm{vis}}^2
\right].
\label{eq:visual-objective}
\end{equation}
Here $\mathcal{D}_k$ contains observation--target pairs for task $k$,
$p(\tau)$ and $w(\tau)$ denote the noise-level sampling distribution and
loss weight, and $\|\cdot\|_{\mathrm{vis}}^2$ denotes the
mean squared velocity-prediction error.
Prompts, noise-level sampling, and loss weighting are specified in
Appendix~\ref{app:perception-prompts}.

\subsection{Joint Training and Inference}
\label{sec:training}
By default, each action forward pass is paired with one perception forward
pass on the same observations. For each sample, a perception task $k$ is drawn
from $p_{\mathcal{K}}$, the sampling distribution over the enabled task set
$\mathcal{K}$. Both losses contribute to the same parameter update through
the joint objective
\begin{equation}
\mathcal{L}
=\mathcal{L}_{\mathrm{act}}
+\lambda_{\mathrm{vis}}\,
\mathbb{E}_{k\sim p_{\mathcal{K}}}
\left[\mathcal{L}_{\mathrm{vis}}^{(k)}\right],
\label{eq:joint-objective}
\end{equation}
where $\lambda_{\mathrm{vis}}\geq0$ weights perception supervision.
Both objectives update the shared backbone; action supervision also
trains the prefix and adapter. The text encoder and VAE remain frozen.

At inference, only the action path in Equation~\ref{eq:action-path} is
executed. GenPA encodes the current observation and predicts an action
chunk with one backbone forward pass and the regression adapter, without
visual generation or iterative action denoising.
The text encoder's instruction embeddings are computed once
and reused while the instruction remains unchanged. VAE encoding and the
backbone forward pass are repeated for each new observation.
Architecture and training details are provided in
Appendix~\ref{app:implementation}; target construction and sampling
are detailed in Appendices~\ref{app:visual-target-encodings}
and~\ref{app:visual-sampling}.

\FloatBarrier
\section{Experiments}
\label{sec:evaluation}

We evaluate whether a pretrained image generator can be directly
adapted into a strong, efficient robot policy without a future visual
prediction objective. Simulation and real-world experiments assess control
performance and robustness, while latency profiling and visual-supervision
ablations examine the sources of inference cost and the effects of
perception supervision.

\subsection{Experimental Setup}
\label{sec:experimental-setup}

We train separate multitask policies on LIBERO and RoboTwin~2.0 without
additional large-scale robot-data pretraining. Unless otherwise stated,
GenPA uses current-depth supervision during training.
Training and evaluation details are provided in
Appendices~\ref{app:implementation} and~\ref{app:baseline-sources}.

\paragraph{LIBERO.}
We train jointly on the Spatial, Object, Goal, and Long suites of
LIBERO~\citep{liu2023libero}, comprising 40 tasks. We replay the 2,000
released demonstrations to construct visual supervision labels, retaining
1,808 successfully replayed demonstrations for training
(Appendix~\ref{app:visual-data}).
{The LIBERO policy configuration uses $224\times224$ views with
training-time image augmentation, whereas the visual-target ablation
configuration retains native $256\times256$ views without augmentation.
Our policies in the external benchmark comparisons and visual-target
ablations use training seed 42.
Internal comparisons under the LIBERO policy configuration report mean
$\pm$ sample standard deviation across three training seeds. LIBERO results
marked \emph{independent} use the same statistics across three evaluation
seeds with new initial-state sets for fixed policies trained with seed 42.}
We evaluate
50 trials per task and report suite success rates and their arithmetic mean.
{The GenPA score in Table~\ref{tab:libero-main} is from checkpoint selection;
selection and independent-evaluation protocols are specified in
Appendix~\ref{app:libero-protocol}.}

\paragraph{LIBERO-Plus.}
We evaluate LIBERO-Plus using the same checkpoint reported on standard
LIBERO, without fine-tuning or separate checkpoint selection. We run one
trial on each of the 10,030 released instances across seven perturbation
categories~\citep{fei2026liberoplus}. {Unless otherwise specified,
we use \emph{Clean-Instruction}, which removes extraneous perturbation
metadata (Appendix~\ref{app:plus-protocol}).} Overall and category success rates
pool outcomes over all instances and within each category, respectively.

\paragraph{RoboTwin 2.0.}
We train one policy jointly across 50 bimanual tasks using 2,500 clean and
25,000 randomized demonstrations with depth annotations
\citep{chen2025robotwin2,yan2026flexpi}. We use a 99\%/1\% episode-level
training/validation split. We evaluate 100 trials
per task under each of the Clean and Randomized conditions and report mean
task success for each condition and their arithmetic mean.
For RoboTwin, we use a wider action
adapter for coordinated bimanual manipulation.

\paragraph{Baselines.}
Our comparisons cover representative VLA and WAM
policies~\citep{kim2025openvlaoft,physicalintelligence2025pi05,
wang2026vlaadapter,luo2026simvla,wu2026vlanext,qwen2026robotmanip,
chen2026lawam,kim2026cosmospolicy,yuan2026fastwam,amap2026abotm05,
zhang2026imagewam,bi2025motus}. Tables indicate additional robot-data
pretraining. LIBERO and RoboTwin baseline scores come from published
reports. For LIBERO-Plus, we re-evaluate six released baselines under
Clean-Instruction using their native inference configurations.

\subsection{Simulation Performance and Robustness}
\label{sec:simulation}

\begin{table}[!t]
\centering
\caption{Performance on LIBERO and model characteristics. Robot PT indicates
additional robot-data pretraining. Bold and underline mark the highest and
second-highest scores in each column, respectively, including ties.}
\label{tab:libero-main}
\vspace{3pt}
\fontsize{8}{9.2}\selectfont
\setlength{\tabcolsep}{2.1pt}
\renewcommand{\arraystretch}{1.08}
\begin{tabular*}{\linewidth}{@{\extracolsep{\fill}}lccccrrrrr@{}}
\toprule
\multirow{2}{*}{Method} & \multirow{2}{*}{\shortstack{Robot\\PT}} & \multicolumn{2}{c}{Parameters} & \multirow{2}{*}{\shortstack{Action\\decoding}} & \multicolumn{5}{c}{LIBERO success (\%)}\\
\cmidrule(lr){3-4}\cmidrule(lr){6-10}
 & & Backbone & Head & & Spatial & Object & Goal & Long & Avg.\\
\midrule
\backbonegroup{10}{TableVLM}{Vision-language backbones}
OpenVLA-OFT$^{j}$ & \robotpt & 7B & 151M & Regr. & 97.7 & 98.0 & 96.1 & 95.3 & 96.80\\
$\pi_{0.5}$ & \robotpt & 3B & 300M & Flow & 98.8 & 98.2 & 98.0 & 92.4 & 96.85\\
VLA-Adapter-Pro & \norobotpt & 1.25B & 218M & Regr. & \underline{99.6} & 99.6 & 98.2 & 96.4 & 98.50\\
SimVLA & \norobotpt & 0.5B & $\sim$300M & Flow & \underline{99.6} & \underline{99.8} & 98.6 & 96.4 & 98.60\\
VLANeXt & \norobotpt & 2B & 612M & Flow & 99.0 & 99.2 & 96.6 & 94.8 & 97.40\\
Qwen-RobotManip & \robotpt & 4B & --- & Flow & --- & --- & --- & --- & \underline{99.10}\\
LaWAM$^{\dagger}$ & \robotpt & 1.52B & 306M & Flow & 99.4 & 99.6 & 98.4 & 97.0 & 98.60\\
\backbonegroup{10}{TableVideo}{Video-generation backbones}
Cosmos Policy & \norobotpt & 2B & Shared & Diffusion & 98.1 & \textbf{100.0} & 98.2 & \underline{97.6} & 98.50\\
Fast-WAM & \norobotpt & 5B & 1B & Flow & 98.2 & \textbf{100.0} & 97.0 & 95.2 & 97.60\\
ABot-M0.5 & \robotpt & 5B & --- & Flow & \textbf{100.0} & \underline{99.8} & \textbf{99.4} & \textbf{98.4} & \textbf{99.40}\\
\backbonegroup{10}{TableImage}{Image-generation backbones}
ImageWAM-4B & \norobotpt & 4B & 642M & Flow & 97.2 & 99.2 & \underline{98.8} & \textbf{98.4} & 98.40\\
\textbf{GenPA} & \norobotpt & 4B & 19M & Regr. & 99.4 & \underline{99.8} & \underline{98.8} & \textbf{98.4} & \underline{99.10}\\
\bottomrule
\end{tabular*}
\vspace{2pt}
\begin{minipage}{\linewidth}\scriptsize
$j$: joint four-suite policy. Regr.: regression. Shared: no separate action expert. ---: unavailable.
$\dagger$: LaWAM additionally uses a latent world decoder and a DINO encoder.
Generative backbone sizes exclude separate text encoders and VAEs.
{Sources and parameter counts: Appendix~\ref{app:libero-protocol}.}
\end{minipage}
\end{table}

\paragraph{LIBERO.}
GenPA achieves \GenPALibero\% mean success on LIBERO
(Table~\ref{tab:libero-main}), the highest average among the listed methods
without additional robot-data pretraining. It comes within
0.30 percentage points of ABot-M0.5, which additionally undergoes
large-scale pretraining on diverse robot datasets. Relative to ImageWAM-4B,
which uses the same image generation
backbone family, GenPA achieves 99.10\% versus 98.40\% mean success
while using a 19M-parameter action adapter rather than a 642M-parameter
action expert. GenPA achieves these results with current-depth
supervision rather than a future visual prediction objective.
{Appendix~\ref{app:libero-validation} provides complete checkpoint
scans and independent LIBERO evaluations of fixed policies on new initial states.}

\begin{table}[!htbp]
\centering
\caption{LIBERO-Plus success under Clean-Instruction evaluation (\%).
All policies are evaluated by us. Bold and underline mark the highest and second-highest scores.}
\label{tab:robust-main}
\vspace{3pt}
\fontsize{8}{9.2}\selectfont
\setlength{\tabcolsep}{2.3pt}
\renewcommand{\arraystretch}{1.05}
\begin{tabular*}{\linewidth}{@{\extracolsep{\fill}}lcrrrrrrrr@{}}
\toprule
Method & \shortstack{Robot\\PT} & Camera & \shortstack{Robot\\Init.} & Lang. & Light & Backgr. & Noise & Layout & Overall\\
\midrule
OpenVLA-OFT & \robotpt & 64.73 & 28.32 & 87.12 & 95.36 & \underline{89.59} & 76.83 & 72.20 & 71.75\\
$\pi_{0.5}$ & \robotpt & 72.23 & \textbf{79.61} & 86.21 & 96.67 & \textbf{96.47} & 89.76 & \underline{86.75} & \textbf{85.90}\\
VLANeXt & \norobotpt & \textbf{89.49} & 62.77 & 79.18 & 95.10 & 89.22 & \textbf{95.57} & 77.90 & 83.60\\
Cosmos Policy & \norobotpt & 80.24 & 64.97 & \underline{88.29} & \textbf{99.12} & 88.85 & 95.07 & \textbf{87.15} & \underline{85.60}\\
Fast-WAM & \norobotpt & 46.40 & \underline{72.39} & 67.86 & 97.11 & 66.91 & 69.52 & 78.30 & 70.22\\
ImageWAM & \norobotpt & 79.36 & 50.19 & \textbf{91.28} & 97.55 & 88.48 & \underline{95.19} & 78.56 & 82.13\\
\textbf{GenPA} & \norobotpt & \underline{81.80} & 67.94 & 84.97 & \underline{97.99} & 82.43 & 92.88 & 81.57 & 83.79\\
\bottomrule
\end{tabular*}
\vspace{2pt}
\begin{minipage}{\linewidth}\scriptsize
Robot PT follows Table~\ref{tab:libero-main}. Released checkpoints and native settings: Appendix~\ref{app:plus-protocol}.
\end{minipage}
\end{table}

\paragraph{LIBERO-Plus.}
GenPA achieves perturbation robustness competitive with the
strongest evaluated policies, without additional robot-data pretraining or
a future visual prediction objective. It reaches \GenPACleanPooled\% success
without fine-tuning (Table~\ref{tab:robust-main}), exceeding the released
ImageWAM policy by 1.66 percentage points and remaining within 2.11 percentage
points of both $\pi_{0.5}$ and Cosmos Policy. GenPA ranks second under camera
and lighting perturbations, while robot initial-state perturbations remain
more challenging. Together with the latency results in
Section~\ref{sec:efficiency}, these findings support direct adaptation of
image generators as an effective approach to robust and efficient robot
control.

\WFclear
\Needspace{14\baselineskip}
\begin{wraptable}[10]{r}{0.49\textwidth}
\vspace{-\intextsep}
\makeatletter\def\@captype{table}\makeatother
\caption{RoboTwin 2.0 success rates (\%).}
\label{tab:robotwin-main}
\centering\fontsize{8}{9.2}\selectfont
\setlength{\tabcolsep}{2pt}
\renewcommand{\arraystretch}{1.05}
\begin{tabular*}{\linewidth}{@{\extracolsep{\fill}}lcrrr@{}}
\toprule
Method & \shortstack{Robot\\PT} & Clean & Rand. & Avg.\\
\midrule
$\pi_{0.5}$ & \robotpt & 82.74 & 76.76 & 79.75\\
Qwen-RobotManip & \robotpt & \textbf{93.4} & 92.5 & 93.0\\
Motus & \robotpt & 88.66 & 87.02 & 87.84\\
Fast-WAM & \norobotpt & 91.88 & 91.78 & 91.83\\
ImageWAM & \norobotpt & 93.20 & \textbf{93.56} & \textbf{93.38}\\
\midrule
\textbf{GenPA} & \norobotpt & \GenPARoboTwinClean & \GenPARoboTwinRandom & \GenPARoboTwinAvg\\
\bottomrule
\end{tabular*}
\par\vspace{2pt}\raggedright\scriptsize
Sources and settings: Appendix~\ref{app:robotwin-protocol}.

\end{wraptable}
\paragraph{RoboTwin 2.0.}
GenPA achieves strong performance across 50 bimanual tasks,
reaching \GenPARoboTwinClean\% and \GenPARoboTwinRandom\% success under Clean
and Randomized conditions, respectively, and averaging \GenPARoboTwinAvg\%
(Table~\ref{tab:robotwin-main}). Its average success is close to Fast-WAM
(91.83\%) and 1.93 percentage points below ImageWAM, the strongest baseline
in this comparison. These results support the applicability of direct
image-generator adaptation to coordinated bimanual manipulation.

\WFclear
\FloatBarrier
\Needspace{18\baselineskip}
\subsection{Real-world Manipulation}
\label{sec:real-world}

\begin{wraptable}[9]{r}{0.49\textwidth}
\vspace{\dimexpr-\intextsep-\baselineskip\relax}
\centering
\caption{USB task success rates (\%).}
\label{tab:usb-draft}
\begingroup
\centering\fontsize{8}{9.2}\selectfont
\setlength{\tabcolsep}{2pt}
\renewcommand{\arraystretch}{1.08}
\begin{tabular*}{\linewidth}{@{\extracolsep{\fill}}lrrrr@{}}
\toprule
Method & Standard & Bg. & Light & Avg.\\
\midrule
$\pi_{0.5}$ & \underline{74.0} & 76.0 & 24.0 & 58.0\\
Fast-WAM & 68.0 & 64.0 & 18.0 & 50.0\\
ImageWAM & \textbf{92.0} & \underline{80.0} & \underline{38.0} & \underline{70.0}\\
\textbf{GenPA} & \textbf{92.0} & \textbf{90.0} & \textbf{52.0} & \textbf{78.0}\\
\bottomrule 
\end{tabular*}
\par\vspace{2pt}\raggedright\scriptsize
Bg./Light: background/lighting perturbations.
Avg.: mean across the three conditions.
\par\endgroup

\end{wraptable}

We evaluate contact-rich USB transfer: one arm unplugs a connector
and inserts it into a fixed target port.
We compare GenPA with $\pi_{0.5}$, Fast-WAM, and ImageWAM,
all fine-tuned on the same 600 demonstrations. GenPA uses current-depth
co-training.
In all three conditions, initial arm pose and source port vary.
\emph{Standard} uses the nominal scene; \emph{Background} adds visual
distractors; and \emph{Lighting} replaces room lighting with a phone LED.
Each policy receives 50 trials per condition (150 total).
Success requires task completion and gripper release within 50 seconds
without dropping the connector.
We report per-condition success and their arithmetic mean;
setup and execution examples appear in Appendix~\ref{app:real-world-details}.

GenPA achieves the highest mean success (78.0\%), matching
ImageWAM under Standard and exceeding it by 10 and 14 percentage points
under Background and Lighting, respectively (Table~\ref{tab:usb-draft}).
These results support direct image-generator adaptation for contact-rich
real-world manipulation under visual perturbations, although lighting
remains challenging.

\WFclear
\FloatBarrier
\Needspace{16\baselineskip}
\subsection{Action Readout and Efficiency}
\label{sec:efficiency}

\newcommand{\GenPAArchRegressionLatency}{27.86}
\newcommand{\GenPAArchFlowLatency}{53.89}
\newcommand{\GenPAArchRegressionCompiledLatency}{20.94}
\newcommand{\GenPAArchFlowCompiledLatency}{29.22}

\paragraph{Action readout.}
We compare GenPA's regression adapter with a comparable flow-based adapter
using the LIBERO {policy} configuration
(Table~\ref{tab:e9-action-readout}; architectural details in
Appendix~\ref{app:action-heads}).

\begin{table}[!htbp]
\centering
\caption{Action readout: success and latency.}
\label{tab:e9-action-readout}
\small
\setlength{\tabcolsep}{5pt}
\renewcommand{\arraystretch}{1.08}
\begin{tabular*}{\linewidth}{@{\extracolsep{\fill}}lrrrr@{}}
\toprule
\multirow{2}{*}{Head} & \multirow{2}{*}{LIBERO (\%)} & \multirow{2}{*}{Plus (\%)} & \multicolumn{2}{c}{Latency (ms)$\downarrow$}\\
\cmidrule(lr){4-5}
 & & & Uncompiled & Compiled\\
\midrule
Flow & {$98.27\pm0.12$} & {$81.49\pm1.18$} & 53.89 & 29.22\\
\textbf{Regression} & {$98.87\pm0.25$} & {$82.33\pm1.27$} & \textbf{27.86} & \textbf{20.94}\\
\bottomrule
\end{tabular*}
\par\vspace{2pt}\raggedright\scriptsize
{Success rates: mean $\pm$ SD across three training runs.}
Latency: H100, cached instructions. Success rates use uncompiled inference;
results after compilation are reported in Appendix~\ref{app:compiled-control}.
\end{table}

Unless otherwise stated, GenPA latency excludes Qwen text encoding, since
instruction embeddings can be reused while the instruction remains unchanged.
{In this comparison, regression achieves higher mean success rates
on both LIBERO and LIBERO-Plus while reducing H100 latency under both
uncompiled and compiled inference. Full comparisons and uncertainty estimates
are provided in Appendix~\ref{app:action-readout-results}.}

\begin{wrapfigure}[17]{r}{0.49\textwidth}
\vspace{-\intextsep}
\centering
\includegraphics[width=\linewidth]{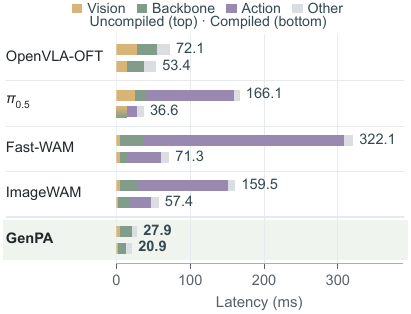}
\caption{\textbf{H100 action-chunk latency.}
Totals are timed separately; Vision and Backbone are jointly profiled
for compiled $\pi_{0.5}$.}
\label{fig:latency-architecture}
\end{wrapfigure}

\paragraph{Latency across policy architectures.}
Figure~\ref{fig:latency-architecture} decomposes latency into visual encoding
(\emph{Vision}), backbone computation (\emph{Backbone}), action decoding
(\emph{Action}), and remaining overhead (\emph{Other});
{Appendix~\ref{app:efficiency} provides timing protocols,
component measurements, and additional latency results.}

\WFclear
\Needspace{20\baselineskip}

Like GenPA, Fast-WAM and ImageWAM reuse independently encoded instructions,
so their timings also exclude text encoding.
The three policies have similar VAE-based visual encoding costs, but
iterative action decoding dominates the uncompiled latency of Fast-WAM
and ImageWAM and remains their largest component after compilation.
GenPA takes 27.9\,ms without compilation and 20.9\,ms with compilation.
The smaller action expert of $\pi_{0.5}$ benefits substantially from
compilation, bringing total latency to 36.6\,ms, below OpenVLA-OFT's
53.4\,ms despite retaining iterative decoding.
OpenVLA-OFT uses regression, but incurs higher visual-encoding and backbone costs.
GenPA thus combines strong manipulation performance with
20.9\,ms action-chunk inference while using a 4B-parameter backbone.
{Deployment memory is reported in Appendix~\ref{app:timing-qwen}.}

\WFclear
\paragraph{Why visual generation backbones suit efficient control.}
A more revealing difference lies in the computation required for each
new observation. Visual encoding is substantially more expensive in the
two evaluated VLM-based policies than VAE encoding in policies built on
visual generation backbones. This reflects different divisions of
computation: VLMs first extract high-level features with a vision encoder
and then perform fusion and reasoning in a multimodal backbone;
latent visual generators use a lightweight VAE to compress observations
into reconstructable latents, perform conditional visual modeling in the
visual generation backbone, and encode language
independently~\citep{rombach2022latentdiffusion}.
Comparing backbone sizes alone therefore overlooks the different
visual-encoding costs these architectures incur at each observation update.
The latter organization matches the inputs of robotic manipulation:
observations change continuously, whereas instructions typically remain
fixed within a task. Language encoding can thus be amortized across policy
calls, while each new observation passes through a lightweight VAE and
the visual generation backbone.
GenPA combines this organization with a lightweight regression adapter,
producing actions through a single backbone forward pass and direct action
readout. These results suggest that efficient policy design should not
only simplify action generation but also align module boundaries with
input update frequencies; visual generation pretraining provides a
favorable architectural basis for this organization.

\WFclear
\FloatBarrier
\Needspace{14\baselineskip}
\subsection{Native Depth Supervision versus Direct Depth Regression}
\label{sec:depth-control}
\begin{wraptable}[9]{r}{0.49\textwidth}\vspace{-\intextsep}
\begingroup
\centering
\caption{Native depth supervision vs. direct depth regression (\%).}
\label{tab:depth-regression-control}
\fontsize{8}{9.5}\selectfont
\setlength{\tabcolsep}{3pt}
\renewcommand{\arraystretch}{1.08}
\begin{tabular*}{\linewidth}{@{\extracolsep{\fill}}lrr@{}}
\toprule
Supervision & LIBERO & Plus \\
\midrule
Native flow & $98.87\pm0.25$ & $82.33\pm1.27$ \\
DPT regression & $98.85\pm0.10$ & $80.61\pm0.35$ \\
\bottomrule
\end{tabular*}
\par\vspace{3pt}
\begin{minipage}{\linewidth}\footnotesize
Mean $\pm$ SD across three training runs.
\end{minipage}
\par\endgroup
\end{wraptable}

To assess the benefit of the native perception interface, we
compare native flow-based depth supervision (Native flow) with direct
depth regression (DPT regression) using a DPT-style head~\citep{ranftl2021dpt}
under the same backbone, depth labels, and action-update budget.
For DPT regression, the depth head and action adapter read the same final reference-token features and
jointly train the backbone in a single forward pass without target-image
tokens. The two approaches perform similarly on LIBERO, while native
depth supervision achieves higher LIBERO-Plus success
(Table~\ref{tab:depth-regression-control}). Training takes 1.294\,s per
update for native depth supervision and 0.517\,s for DPT regression on eight
H100 GPUs. Implementation details and full results are provided in
Appendices~\ref{app:auxiliary-depth-head} and~\ref{app:direct-depth-control};
training resource costs are summarized in Appendix~\ref{app:training-cost}.

\WFclear
\FloatBarrier
\subsection{Visual Supervision and Design Choices}
\label{sec:analysis}

We examine whether current-scene perception can provide effective
supervision without future prediction, and how target choice and combination
affect robustness. {To limit preprocessing-related confounds when
comparing supervision targets, Table~\ref{tab:visual-supervision} uses the
visual-target ablation configuration. All configurations in this table share
the same model architecture, training data, and maximum action-update budget.}
Target encodings and sampling details are given in
Appendices~\ref{app:visual-target-encodings}
and~\ref{app:visual-sampling}.

\begin{table}[!htbp]
\centering
\caption{Effect of visual supervision on closed-loop success (\%).
LIBERO and Plus give absolute rates; perturbations give None rates and
changes from None (percentage points).}
\label{tab:visual-supervision}
\fontsize{8}{9}\selectfont
\setlength{\tabcolsep}{2pt}
\renewcommand{\arraystretch}{0.95}
\setlength{\aboverulesep}{0.25ex}
\setlength{\belowrulesep}{0.35ex}
\begin{tabular*}{\linewidth}{@{\extracolsep{\fill}}l@{\hspace{6pt}}rr@{\hspace{5pt}}rrrrrrr@{}}
\toprule
Target & {\shortstack{LIBERO\\(independent)}} & Plus & Camera & \shortstack{Robot\\Init.} & Lang. & Light & Bg. & Noise & Layout\\
\midrule
\rowcolor{black!6}
None & {$97.02\pm0.15$} & 70.87 & 66.42 & 59.10 & 79.38 & 93.08 & 43.40 & 73.52 & 78.89 \\
D & {$97.87\pm0.65$} & 76.69 & \cellcolor{GenPABlue!17.03}$+7.6$ & \cellcolor{GenPABlue!17.27}$+7.7$ & \cellcolor{GenPABlue!18.01}$+8.0$ & \cellcolor{GenPABlue!6.50}$+2.9$ & \cellcolor{GenPABlue!3.97}$+1.8$ & \cellcolor{GenPABlue!18.41}$+8.2$ & \cellcolor{GenPABlue!5.61}$+2.5$ \\
N & {$96.52\pm0.53$} & 71.70 & \cellcolor{GenPAOrange!3.52}$-1.6$ & \cellcolor{GenPABlue!14.81}$+6.6$ & \cellcolor{GenPABlue!5.42}$+2.4$ & \cellcolor{GenPAOrange!10.64}$-4.7$ & \cellcolor{GenPABlue!19.87}$+8.8$ & \cellcolor{GenPAOrange!6.46}$-2.9$ & \cellcolor{GenPAOrange!3.69}$-1.6$ \\
S & {$97.17\pm0.39$} & 76.08 & \cellcolor{GenPABlue!2.67}$+1.2$ & \cellcolor{GenPABlue!12.63}$+5.6$ & \cellcolor{GenPABlue!8.64}$+3.8$ & \cellcolor{GenPABlue!6.50}$+2.9$ & \cellcolor{GenPABlue!33.67}$+15.0$ & \cellcolor{GenPABlue!23.61}$+10.5$ & \cellcolor{GenPAOrange!0.59}$-0.3$ \\
P & {$97.52\pm0.60$} & 72.19 & \cellcolor{GenPAOrange!10.55}$-4.7$ & \cellcolor{GenPABlue!0.87}$+0.4$ & \cellcolor{GenPABlue!12.44}$+5.5$ & \cellcolor{GenPABlue!5.91}$+2.6$ & \cellcolor{GenPABlue!30.32}$+13.5$ & \cellcolor{GenPAOrange!6.61}$-2.9$ & \cellcolor{GenPAOrange!1.62}$-0.7$ \\
G & {$96.62\pm0.16$} & 71.80 & \cellcolor{GenPAOrange!16.32}$-7.3$ & \cellcolor{GenPABlue!0.00}$+0.0$ & \cellcolor{GenPABlue!15.37}$+6.8$ & \cellcolor{GenPABlue!2.36}$+1.1$ & \cellcolor{GenPABlue!8.78}$+3.9$ & \cellcolor{GenPABlue!5.20}$+2.3$ & \cellcolor{GenPABlue!2.07}$+0.9$ \\
\midrule
F & {$96.98\pm0.20$} & 77.10 & \cellcolor{GenPABlue!18.29}$+8.1$ & \cellcolor{GenPABlue!10.31}$+4.6$ & \cellcolor{GenPABlue!3.81}$+1.7$ & \cellcolor{GenPABlue!3.15}$+1.4$ & \cellcolor{GenPABlue!40.36}$+17.9$ & \cellcolor{GenPABlue!27.83}$+12.4$ & \cellcolor{GenPAOrange!1.33}$-0.6$ \\
FD & {$98.02\pm0.20$} & 74.46 & \cellcolor{GenPAOrange!9.99}$-4.4$ & \cellcolor{GenPABlue!16.98}$+7.5$ & \cellcolor{GenPABlue!11.42}$+5.1$ & \cellcolor{GenPAOrange!6.70}$-3.0$ & \cellcolor{GenPABlue!43.91}$+19.5$ & \cellcolor{GenPABlue!3.79}$+1.7$ & \cellcolor{GenPABlue!4.87}$+2.2$ \\
\midrule
D+S & {$97.23\pm0.53$} & 75.09 & \cellcolor{GenPABlue!15.34}$+6.8$ & \cellcolor{GenPABlue!14.37}$+6.4$ & \cellcolor{GenPABlue!1.32}$+0.6$ & \cellcolor{GenPABlue!5.32}$+2.4$ & \cellcolor{GenPABlue!22.17}$+9.9$ & \cellcolor{GenPABlue!6.32}$+2.8$ & \cellcolor{GenPABlue!4.28}$+1.9$ \\
D+P & {$97.78\pm0.28$} & 75.53 & \cellcolor{GenPABlue!24.48}$+10.9$ & \cellcolor{GenPABlue!8.13}$+3.6$ & \cellcolor{GenPABlue!11.42}$+5.1$ & \cellcolor{GenPABlue!3.55}$+1.6$ & \cellcolor{GenPABlue!6.69}$+3.0$ & \cellcolor{GenPABlue!14.62}$+6.5$ & \cellcolor{GenPABlue!0.89}$+0.4$ \\
D+F & {$97.37\pm0.47$} & 77.42 & \cellcolor{GenPABlue!13.65}$+6.1$ & \cellcolor{GenPABlue!20.76}$+9.2$ & \cellcolor{GenPABlue!11.13}$+4.9$ & \cellcolor{GenPABlue!0.59}$+0.3$ & \cellcolor{GenPABlue!27.60}$+12.3$ & \cellcolor{GenPABlue!26.28}$+11.7$ & \cellcolor{GenPABlue!2.80}$+1.2$ \\
D+N+S+P$^{\dagger}$ & {$97.77\pm0.33$} & 75.45 & \cellcolor{GenPABlue!21.11}$+9.4$ & \cellcolor{GenPABlue!1.74}$+0.8$ & \cellcolor{GenPABlue!1.90}$+0.8$ & \cellcolor{GenPABlue!3.15}$+1.4$ & \cellcolor{GenPABlue!9.62}$+4.3$ & \cellcolor{GenPABlue!27.97}$+12.4$ & \cellcolor{GenPABlue!3.54}$+1.6$ \\
\bottomrule
\end{tabular*}
\par\vspace{2pt}
\begin{minipage}{\linewidth}\scriptsize
{Each row uses one fixed checkpoint for both benchmarks; LIBERO reports mean $\pm$ SD across three evaluation seeds.}
D: current depth; N: normals; S: segmentation; P: pointmaps; G: task-conditioned grounding;
F: future RGB; FD: future depth.
$^{\dagger}$Two perception passes per action pass, with summed losses.
All other configurations use one, except None, which omits perception supervision.
\end{minipage}
\end{table}

\paragraph{Current perception provides effective control supervision.}
Current depth (D) raises LIBERO-Plus success from 70.87\% to 76.69\%.
Future depth (FD), using the same depth encoding for a target 16 retained
action steps into the future,
reaches a similar {independent LIBERO score (98.02\% versus 97.87\%)} but lower
LIBERO-Plus success (74.46\%). Future RGB (F) remains effective at
77.10\%, despite lower {independent LIBERO success (96.98\%)}.
Standard-task success alone is therefore insufficient for
choosing an auxiliary objective. These results suggest that future prediction
is not a prerequisite for improving closed-loop robustness through visual
supervision.
{Under the LIBERO policy configuration, mean LIBERO-Plus success increases from $79.49\pm0.85$\%
without visual supervision to $82.33\pm1.27$\% with current depth and
$81.80\pm2.05$\% with future RGB
(Appendix~\ref{app:reported-config-supervision}).}

\paragraph{More targets do not consistently improve control.}
Although D and S each improve robustness, D+S scores below
either alone. D+P improves camera robustness over D but lowers overall
success, showing that category-specific gains need not improve robustness
overall. The four-target configuration also scores below D despite using
more perception computation. D+F achieves the highest overall score, but
exceeds F by only 0.32 percentage points. These results suggest that
informative visual targets are not automatically complementary for control:
their joint benefit must be established through closed-loop evaluation.
Current-depth supervision remains a simple, effective default.

\paragraph{Why is depth supervision effective?}
Depth improves success over None in all seven perturbation
categories, whereas other geometric targets do not yield equally broad gains.
We hypothesize that depth benefits from combining control-relevant geometry
with a simple supervision format. It provides metric distance constraints
on visible surfaces, encouraging the backbone to learn spatial structure
relevant to reaching, grasping, and placement. This information is expressed
as a dense scalar field aligned with the current observation, whose definition
does not depend on task roles or execution stages. In contrast, robot-frame
pointmaps additionally require learning a mapping from the current view to
the robot frame; normals primarily constrain surface orientation rather than
metric distance; and segmentation emphasizes categories and boundaries.
These differences suggest that the value of auxiliary supervision depends
not only on its information content, but also on what it asks the backbone
to learn and how these constraints support control.

\section{Related Work}
\label{sec:related_work}

\paragraph{Visual generative pretraining for robot control.}
Pretrained image and video generators provide visual subgoals for
goal-conditioned policies~\citep{black2023susie} or predicted transitions
for inverse dynamics~\citep{li2026lingbotva}. Other methods jointly generate
future observations and actions~\citep{shen2025videovla,kim2026cosmospolicy,ye2026dreamzero},
or retain future prediction for co-training while enabling action-only
inference~\citep{yuan2026fastwam,ye2026gigaworldpolicy}. The closely related
ImageWAM combines image-editing pretraining, future-image co-training, and
a flow-based action expert, supporting inference without image generation
while retaining iterative action decoding~\citep{zhang2026imagewam}.
Beyond future prediction, generators can represent actions through
joint-localization targets, pixel-motion fields, or multiview action
images~\citep{shridhar2024genima,nguyen2026dawn,zhen2026actionimages}.
ORCA instead extracts task-adaptive features from a frozen diffusion model
through learned conditioning~\citep{shin2026orca}.
{Whereas Fast-WAM and ImageWAM retain future-prediction co-training,}
GenPA fine-tunes the image generation backbone for direct
action prediction through a lightweight regression adapter, coupled with
current-scene perception co-training.

\paragraph{Perceptual supervision for policy adaptation.}
Auxiliary perception objectives have been studied across policy
architectures. They range from future visual
prediction~\citep{wu2023gr1,tian2025seer,zhang2025dreamvla} to current-scene
depth supervision~\citep{li2025qdepth,yuan2025depthvla} and localization,
segmentation, or spatial-relation tasks~\citep{tu2026sgvla,li2024llara}.
Geometric pretraining and supervision further support policy
adaptation~\citep{song2026vga,peng2026g3vla}, while representation analyses
link state decodability to control performance~\citep{dong2026capturing}.
LingBot-VLA~2.0 distills depth and video representations at
both current and future timesteps~\citep{wu2026lingbotvla2}. Generative perception research
demonstrates that visual generators can learn dense
geometry~\citep{ke2024marigold,he2025lotus} and diverse instruction-conditioned
perception tasks~\citep{gabeur2026visionbanana,han2026sensenovavision,wang2026genception}.
GenPA uses the generator's native image-editing interface for
perception co-training without task-specific perception heads.
Perceptual objectives shape the shared backbone during training; predicted
annotations are not inputs to the deployed policy.
Our ablations compare current-scene and future targets, as well
as target combinations, within a common policy architecture to assess their
utility for closed-loop control.
{We further compare native depth supervision with a direct
depth-regression head to evaluate their control performance and training cost.}

\paragraph{Action prediction and inference efficiency.}
Action chunking is established in visuomotor imitation~\citep{zhao2023act},
with actions produced through diffusion or flow
matching~\citep{chi2023diffusion,black2025pi0}, autoregressive token
decoding~\citep{pertsch2025fast}, or parallel regression~\citep{kim2025openvlaoft}.
Lightweight action adapters~\citep{wang2026vlaadapter}, fewer generative
decoding steps~\citep{guo2026reactvla}, and decoupling action-module depth
from the visual backbone~\citep{ma2026fasterwamdot} offer different routes
to efficient inference. GenPA combines lightweight regression with an
adapted image generation backbone and {compares regression and
flow-based action readout under a common training framework to examine whether
strong closed-loop control requires iterative action denoising.} Beyond action decoding,
our component-level analysis examines visual encoding and backbone
computation, relating architectural differences to the recurring costs
of changing observations and the reusable computation associated with
fixed instructions.

\FloatBarrier
\section{Conclusion}
\label{sec:conclusion}
We presented GenPA, which directly adapts a pretrained image generation
model into a robot policy through lightweight action regression and native
perception co-training, without requiring future visual prediction as an
intermediate training objective. At deployment, GenPA requires only a single
backbone forward pass and action regression.
GenPA combines strong control performance and perturbation robustness
in simulation and real-world manipulation with low measured inference latency. Ablations of visual
supervision show that depth supervision on current observations offers a
simple, effective training recipe, while adding targets does not always
improve control. Our latency analysis further suggests that efficient policy
design should not only simplify action decoding but also match computation
to input update frequencies. Visual generation models provide a favorable
architectural basis for this organization: a lightweight VAE encodes changing
observations, while independently encoded language features can be reused
across observations.

\label{page:main-end}

\clearpage
\section*{AI Use Statement}
We used generative AI tools, including ChatGPT and Codex,
to assist with writing, translation, and language polishing; literature
retrieval and discovery; and research execution, including
experimental design, implementation and debugging, data processing, and
result interpretation. GPT Image 2 was used to produce the illustrative
image edits in Figure~\ref{fig:genpa-overview}(a); the source image and
complete prompts are documented in Appendix~\ref{app:figure1-generation}.
The authors reviewed and revised the AI-assisted material and take full
responsibility for the final manuscript, methodology, and reported results.

\section*{Reproducibility Statement}
Appendix~\ref{app:implementation} provides the model
architecture, input and output interfaces, preprocessing, and training
configurations. Appendix~\ref{app:visual-supervision} describes visual-target
construction and the visual-supervision ablation protocol.
Appendix~\ref{app:baseline-sources} documents simulation evaluation,
checkpoint selection, baseline configurations, and LIBERO-Plus instruction
processing. {Appendix~\ref{app:efficiency} provides timing
protocols, detailed latency measurements, and closed-loop evaluation after
compilation. The real-world setup and evaluation protocol are described in
Appendix~\ref{app:real-world-details}.} For code, model checkpoints, and
data, see the \href{\projecturl}{GenPA project page}.

\bibliography{references}
\bibliographystyle{iclr2027_conference}
\clearpage
\appendix
\noindent
This appendix provides implementation and training details
(Appendix~\ref{app:implementation}), visual supervision and ablation details
(Appendix~\ref{app:visual-supervision}), simulation evaluation protocols {and validation}
(Appendix~\ref{app:baseline-sources}), {action-readout comparisons and
measurements of inference latency and memory}
(Appendix~\ref{app:efficiency}), real-world experimental details
(Appendix~\ref{app:real-world-details}), and generation details for Figure~\ref{fig:genpa-overview}
(Appendix~\ref{app:figure1-generation}).

\section{Implementation and Training Details}
\label{app:implementation}

This appendix specifies GenPA's model architecture, tensor interfaces,
image preprocessing, and training configuration. Perception-target
construction {and auxiliary-objective details}
are provided in Appendix~\ref{app:visual-supervision};
{simulation protocols and checkpoint selection} in
Appendix~\ref{app:baseline-sources}; and platform-specific real-world
implementation in Appendix~\ref{app:real-world-details}.

\subsection{Model Architecture and Trainable Components}
\label{app:model-architecture}

GenPA is initialized from FLUX.2 klein Base 4B
\citep{blackforestlabs2026flux2klein}.
Table~\ref{tab:appendix-model-components} summarizes the architecture
and training status of each component. Following the native FLUX.2 conditioning
pipeline, we concatenate the frozen Qwen3-4B encoder's hidden states from
layers 9, 18, and 27 at each text position to produce 7,680-dimensional
conditioning features.

The frozen VAE encodes observations and perception targets into
32-channel latent feature maps. Packing each $2\times2$ latent neighborhood
produces 128-dimensional image tokens. The VAE decoder is not used in
either the action forward pass or the perception forward pass.

All DiT parameters are fine-tuned. {The action forward pass
contains no target-image tokens and uses the final reference-token features
for action prediction. In the perception forward pass, the native output
layer maps the final target-token features to latent velocities and receives
gradients only from perception supervision.} The Action Prefix and action
adapter are randomly initialized.

{With instruction embeddings cached and Qwen unloaded, the LIBERO
policy contains approximately 3.98B parameters, including the frozen VAE;
keeping Qwen resident increases this to 8.00B. The 19M figure refers only
to the action adapter.}

\begin{table}[htbp]
\centering
\caption{GenPA components and their training status.}
\label{tab:appendix-model-components}
\footnotesize
\setlength{\tabcolsep}{3.5pt}
\renewcommand{\arraystretch}{1.12}
\hyphenpenalty=10000
\exhyphenpenalty=10000
\begin{tabularx}{\linewidth}{@{}>{\raggedright\arraybackslash}p{0.17\linewidth}>{\raggedright\arraybackslash}p{0.37\linewidth}>{\raggedright\arraybackslash}X>{\raggedright\arraybackslash}p{0.13\linewidth}@{}}
\toprule
Component & Architecture & Role & Status \\
\midrule
Qwen3-4B & 36 layers, width 2,560; 32 attention heads, 8 key--value heads &
Text conditioning for both forward passes & Frozen \\
FLUX.2 VAE & 4 stages, widths 128/256/512/512; 2 residual blocks per stage; $8\times$ spatial compression &
Encode observations and perception targets & Frozen \\
Visual generation DiT & 5 dual-stream + 20 single-stream blocks; width 3,072; 24 heads of dimension 128; MLP ratio 3 &
Shared action--perception backbone & Trainable \\
Native output layer & Non-affine LayerNorm, timestep modulation, and bias-free $3{,}072\!\rightarrow\!128$ projection inside the DiT &
Perception-flow prediction & Trainable \\
Action Prefix & 16 learned 7,680-dimensional tokens &
Action-specific backbone conditioning & Trainable \\
Action adapter & Four-layer Transformer decoder; see Appendix~\ref{app:action-heads} &
Action-chunk prediction & Trainable \\
\bottomrule
\end{tabularx}
\end{table}

The action and perception forward passes use the token groups, attention
patterns, and time conditioning defined in
Sections~\ref{sec:action-path} and~\ref{sec:visual-supervision}, respectively.

\subsection{Action Adapters}
\label{app:action-heads}

\paragraph{Regression action adapter.}
The default adapter layer-normalizes the final reference-token states and
projects them from 3,072 to 512 dimensions. We add two-dimensional sinusoidal
position features derived from the image-token coordinates and apply a second
layer normalization to form the visual memory. Proprioception is encoded by
LayerNorm--Linear--GELU--Linear, with both linear outputs of width 512, and
added to each of 16 learned action queries in the simulation configurations.
Four pre-normalized decoder blocks
apply bidirectional self-attention among the queries, cross-attention to the
visual memory, and a feed-forward network. Each attention layer has eight
heads; the feed-forward width is 2,048, with GELU activation and zero dropout.
A final LayerNorm and linear projection predict the complete action chunk in
parallel. The adapter reads the backbone output without writing states back
into the backbone.

For RoboTwin~2.0, we retain the four-layer, eight-head architecture and 16
action queries, but increase the adapter width to 1,024 and the feed-forward
width to 4,096. The visual-memory projection, proprioception encoder, learned
queries, and output projection use this wider representation.

\paragraph{Flow-based action adapter.}
{The flow-based adapter matches the default regression adapter in}
backbone, Action Prefix, visual-memory projection, decoder depth, hidden
width, and attention-head count. {Each noised seven-dimensional
action is linearly projected to width 512, replacing the learned action
queries. Proprioception is encoded by the same MLP as in the regression
adapter, followed by LayerNorm, and appended as one token to the visual
memory.} Self-attention uses one-dimensional RoPE {with base
10,000} over the 16 action positions.
Flow time is represented by a 256-dimensional sinusoidal embedding
of $1000\tau$, followed by a two-layer width-512 MLP with SiLU. Each decoder
block uses the embedding to modulate its three pre-normalized sublayers; the
modulation projections are zero initialized. {A final LayerNorm
and linear projection predict the seven-dimensional action velocity.}

For a normalized demonstration chunk $\mathbf{A}_t$, we sample one time
$\tau\sim\mathcal{U}(0,1)$ per chunk and Gaussian noise
$\boldsymbol\epsilon$ of the same shape. {Following the flow
convention in Equation~\ref{eq:visual-flow}, the noised action chunk is}
\begingroup
\begin{equation}
\mathbf{A}_{t,\tau}=(1-\tau)\mathbf{A}_t+\tau\boldsymbol\epsilon.
\label{eq:appendix-action-flow}
\end{equation}
\endgroup
{The adapter predicts the velocity
$\boldsymbol\epsilon-\mathbf{A}_t$ and minimizes mean-squared error over valid
action elements within each chunk, averaged across the batch.
Key padding masks prevent attention to padded action steps, which are also
excluded from the loss.} At inference, eight uniform Euler steps integrate
from $\tau=1$ to zero. The VAE
and DiT execute once per observation, and the decoder's cross-attention keys
and values are cached throughout the action loop. The action-flow time is
independent of the perception-flow time; {the backbone retains
the fixed condition $\tau_A=0$ during the action forward pass.}

\begin{table}[htbp]
\centering
\caption{Action-adapter configurations. Parameter counts include the complete
adapter, including visual-memory and proprioception projections, but exclude
the Action Prefix.}
\label{tab:appendix-action-adapters}
\footnotesize
\setlength{\tabcolsep}{4.5pt}
\begin{tabular}{lrrrrrr}
\toprule
Configuration & Width & FFN & Layers & Heads & Action dim. & Parameters \\
\midrule
LIBERO regression & 512 & 2,048 & 4 & 8 & 7 & 18.68M \\
LIBERO flow-based & 512 & 2,048 & 4 & 8 & 7 & 25.38M \\
RoboTwin 2.0 regression & 1,024 & 4,096 & 4 & 8 & 14 & 71.45M \\
\bottomrule
\end{tabular}
\end{table}

\subsection{Input and Output Interfaces}
\label{app:input-output-interfaces}

\paragraph{{Text conditioning.}}
The text-conditioning sequence contains 128 positions in both forward
passes. In the action forward pass, 16 learned Action Prefix tokens occupy
the first positions, leaving up to 112 positions for instruction embeddings.
In the perception forward pass, no Action Prefix is inserted, so all 128
positions are available for the perception prompt. Unused positions are
right padded and masked.

\paragraph{{Visual inputs and perception outputs.}}
For an image canvas of height $H_I$ and width $W_I$, the VAE produces a
latent tensor in $\mathbb{R}^{32\times H_I/8\times W_I/8}$. Packing
$2\times2$ neighborhoods gives $(H_I/16)(W_I/16)$ image tokens of dimension
128. Reference tokens encode the current observations. Target tokens have
the same layout but are instantiated only in the perception forward pass.
The perception output is a velocity tensor aligned with these target tokens.

\paragraph{{Proprioception and actions.}}
\label{app:action-normalization}
{The simulation action output is a chunk in
$\mathbb{R}^{16\times d_a}$, where $d_a$ is the action dimension.}
For LIBERO, we retain the dataset's seven-dimensional control action and
eight-dimensional proprioception. {Both use dimension-wise
min--max normalization to $[-1,1]$.}

For RoboTwin~2.0, the policy predicts 12 absolute target joint angles and two
continuous gripper commands. Proprioception contains the corresponding
14-dimensional current state. {Both joint and gripper commands
use dimension-wise z-score normalization, while proprioception uses min--max
normalization.}

Table~\ref{tab:appendix-input-interfaces} lists the simulation interfaces.
The flow-based action comparison uses the same interface as the LIBERO
policy. Image sizes are height $\times$ width, and the execution horizon is
the number of predicted actions applied before the next observation.

\begin{table}[htbp]
\centering
\caption{Simulation input and output interfaces. The ``Image tokens'' column reports the number of reference tokens. Perception passes use the same number of target tokens. Pred./exec. denotes predicted/executed actions per chunk.}
\label{tab:appendix-input-interfaces}
\footnotesize
\setlength{\tabcolsep}{4pt}
\begin{tabularx}{\linewidth}{@{}>{\raggedright\arraybackslash}Xrrrrrr@{}}
\toprule
Configuration & Views & RGB canvas & Image tokens & State dim. & Action dim. & Pred./exec. \\
\midrule
LIBERO policy & 2 & $224\times448$ & 392 & 8 & 7 & 16/12 \\
RoboTwin 2.0 policy & 3 & $288\times256$ & 288 & 14 & 14 & 16/16 \\
{Visual-target ablations} & 2 & $256\times512$ & 512 & 8 & 7 & 16/12 \\
\bottomrule
\end{tabularx}
\end{table}

\subsection{Image Preprocessing and Augmentation}
\label{app:image-preprocessing}

{Both the action and perception forward passes use the same
preprocessed RGB observations; the perception forward pass additionally uses
the corresponding perception targets. For depth supervision, observations
and targets share spatial preprocessing and geometric augmentation, while
photometric augmentation is applied only to RGB observations.}

\paragraph{{Augmentation.}}
{Augmentation is applied only during training.
For the LIBERO and RoboTwin 2.0 policies, each training sample
is augmented with probability 0.65. When enabled, each camera independently
selects additive noise, color adjustments, or both with probabilities
$1/4$, $1/4$, and $1/2$, respectively, followed by geometric transforms.
Table~\ref{tab:appendix-augmentation} lists the operations and parameters
in each group. Evaluation uses deterministic preprocessing only.}

\begin{table}[htbp]
\centering
\caption{{Training-time image augmentation.}}
\label{tab:appendix-augmentation}
\footnotesize
\setlength{\tabcolsep}{4pt}
\renewcommand{\arraystretch}{1.15}
\begingroup
\hyphenpenalty=10000
\exhyphenpenalty=10000
\begin{tabular*}{\linewidth}{@{\extracolsep{\fill}}llcl@{}}
\toprule
Group & Operation & Value / range & Applied to \\
\midrule
\multirow{6}{*}{Color adjustments} & Brightness factor & $[0.7,1.3]$ &
\multirow{7}{*}{RGB observations} \\
& Contrast factor & $[0.7,1.3]$ & \\
& Saturation factor & $[0.75,1.25]$ & \\
& Hue shift & $[-0.04,0.04]$ & \\
& Gamma & $[0.8,1.25]$ & \\
& Exposure & $[-0.25,0.25]$ EV & \\
\addlinespace[2pt]
Additive noise & Gaussian noise (std.) & 0.015 & \\
\midrule
\multirow{2}{*}{Geometric transforms} & Crop (side-length ratio) & $[0.92,1]$ &
\multirow{2}{*}{\shortstack[l]{RGB observations\\and depth targets}} \\
& Rotation & $[-8^\circ,8^\circ]$ & \\
\bottomrule
\end{tabular*}
\par\smallskip
\begin{minipage}{\linewidth}
Parameters given as intervals are sampled uniformly. Noise standard deviation
is specified on a $[0,1]$ intensity scale. Crops preserve aspect ratio and
are resized back to the pre-crop resolution. Resizing uses bilinear
interpolation with antialiasing. Rotation uses
bilinear interpolation, with mean RGB intensity as fill for observations
and zero-valued pixels for depth targets. No horizontal flips are applied.
\end{minipage}
\endgroup
\end{table}

\paragraph{LIBERO policy.}
{Each $256\times256$ camera image is first resized to
$224\times224$, augmented during training, and then concatenated
horizontally with the other view.} The regression and flow-based policies
in the action-readout comparison use the same processing.

\paragraph{RoboTwin 2.0 policy.}
Each of the three camera images is first resized to $240\times320$.
{After augmentation during training,} the main view is
resized to $192\times256$
and the two wrist views to $96\times128$ each.
The main view is placed above the side-by-side wrist views, producing a
$288\times256$ canvas.

\paragraph{{Visual-target ablations.}}
These experiments retain both native $256\times256$ views, concatenate them
into a $256\times512$ canvas, and apply no stochastic augmentation or spatial
resizing. All perception targets and corresponding RGB observations follow
the same image preprocessing.

\subsection{Training and Optimization}
\label{app:training-recipes}

{We train all GenPA policies reported in this paper with RADAR,
an optimizer from the RAD family~\citep{lyu2025conformal,ren2026momentum}.} We set
$(\beta_1,\beta_2)=(0.9,0.95)$, enable decoupled weight decay with coefficient
0.01, and use $\gamma=0$, $l=0$, $\delta=1$, and
$\zeta=10^{-16}$.

{During {native perception} co-training, action and perception losses are
backpropagated separately, with gradients accumulated before each optimizer
update.} Configurations that sample multiple perception targets are specified in
Appendix~\ref{app:visual-sampling}. We train in BF16 and clip the gradient norm at 1.0. {Seed settings are specified in Appendix~\ref{app:libero-protocol}.}
Learning rates warm up linearly over the first 5\% of
updates and then follow cosine decay to an absolute minimum of $10^{-6}$ for
each parameter group. The action-interface learning rate applies to both the
Action Prefix and adapter.
{The training configuration uses eight NVIDIA H100 GPUs
with 80\,GB of memory each.}

{The LIBERO policy configuration in the first column also applies
to the regression--flow action-readout comparison and the native-depth--DPT
supervision comparison.}

\begin{table}[htbp]
\centering
\caption{Simulation training configurations for GenPA.}
\label{tab:appendix-training-recipes}
\footnotesize
\setlength{\tabcolsep}{3pt}
\begin{tabularx}{\linewidth}{@{}lXXX@{}}
\toprule
Hyperparameter & {\shortstack{LIBERO policy and\\comparisons}} &
{\shortstack{LIBERO visual-target\\ablations}} & RoboTwin 2.0 policy \\
\midrule
Peak backbone LR & $10^{-5}$ & $10^{-5}$ & $10^{-4}$ \\
Peak action-interface LR & $10^{-4}$ & $10^{-4}$ & $10^{-4}$ \\
Global batch size & 80 & 80 & 384 \\
Maximum updates & 36,750 & 36,750 & 70,847 \\
Training epochs & 10 & 10 & 5 \\
Perception coefficient & 0.1 & 0.1 per sampled target & 0.1 \\
\bottomrule
\end{tabularx}
\end{table}

\begingroup

\paragraph{Training cost.}
\label{app:training-cost}
Table~\ref{tab:training-resources} reports resource costs under the LIBERO
policy configuration, using BF16 and ZeRO-1 with global batch 80.
Update times are averaged over three runs; memory is the maximum per-GPU
PyTorch allocation during the training loop.
GPU-hours are estimated from measured update times for 36,750 updates,
excluding initialization and checkpoint saving.

\begin{table}[htbp]
\centering

\caption{Training resources on eight H100 80GB GPUs.}
\label{tab:training-resources}
\footnotesize
\setlength{\tabcolsep}{5pt}
\begin{tabular*}{\linewidth}{@{\extracolsep{\fill}}lrrr@{}}
\toprule
Training & Update time (s) & Peak memory (GiB/GPU) & Estimated GPU-hours \\
\midrule
Action-only & 0.501 & 35.43 & 40.9 \\
Native depth & 1.294 & 54.11 & 105.7 \\
\bottomrule
\end{tabular*}
\end{table}
\par
\endgroup

\begingroup

\subsection{Auxiliary Depth-Regression Head}
\label{app:auxiliary-depth-head}

For the direct-depth comparison in Section~\ref{sec:depth-control}, we use
a randomly initialized DPT-style decoder~\citep{ranftl2021dpt}
with 11.064M parameters, branch widths of [128, 256, 512, 512], and a
fusion width of 128. All four branches read the same final reference-token
features. The two camera views are decoded separately with shared weights
to produce two $224\times224$ depth maps. We add two-dimensional sinusoidal
positional embeddings after branch projection and final upsampling, and
use an unconstrained linear output.

The depth head and action adapter share a single backbone forward pass
without target-image tokens. We jointly optimize the action loss and a
pixel-wise depth MSE loss with coefficient 0.1, using encoded depth targets
in $[0,1]$. Both losses update the shared backbone. The depth head is
omitted at inference.
\par
\endgroup

\FloatBarrier
\section{Visual Supervision and Ablation Details}
\label{app:visual-supervision}

This appendix specifies the image-form targets in
Section~\ref{sec:visual-supervision} and the protocol behind
Table~\ref{tab:visual-supervision}. Model interfaces and image
processing are specified in Appendix~\ref{app:implementation};
reported-policy and baseline simulation protocols are specified in
Appendix~\ref{app:baseline-sources}.
The visual targets supervise the perception path during training
and are not inputs to the deployed action path.

\subsection{Visual Tasks as Image Targets}
\label{app:visual-target-encodings}

\begin{figure}[!t]
\centering
\includegraphics[width=.90\linewidth]{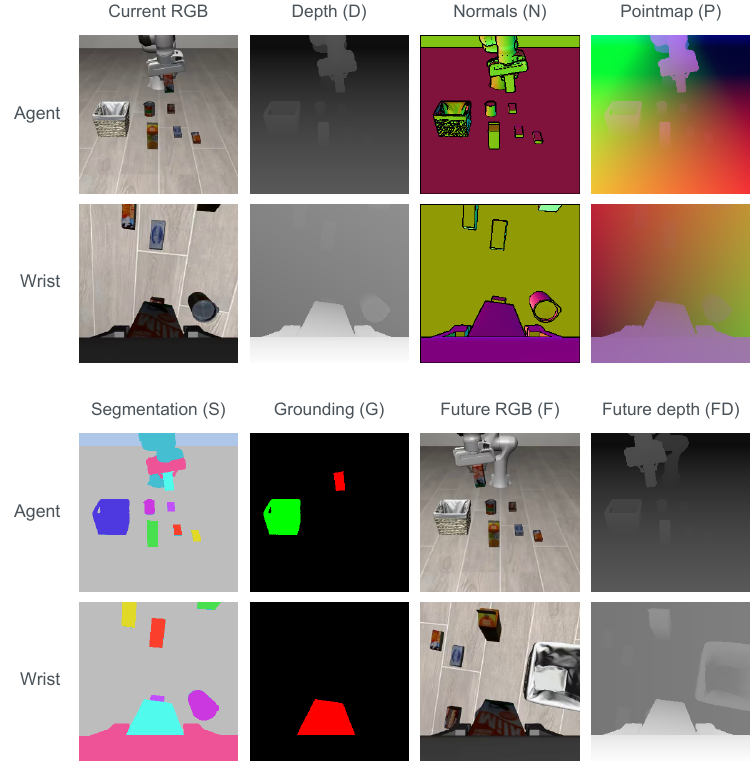}
\caption{Aligned LIBERO observations and supervision targets in agent and wrist views. D, N, P, S, and G denote current depth, normals, pointmaps, segmentation, and grounding. F and FD use RGB and depth from the same future frame. All panels show training data, not model predictions.}
\label{fig:appendix-visual-targets}
\end{figure}

\paragraph{{Image representations.}}
{Each visual target is represented as a three-channel image
paired with the current observation, with values in $[0,1]$.}
{LIBERO D and G use floating-point image values, whereas
N, P, and S use fixed 8-bit RGB encodings.}
F uses the future RGB image, and FD follows D's floating-point encoding.
Figure~\ref{fig:appendix-visual-targets} shows the current two-view
RGB reference alongside its current- and future-state targets.

\paragraph{Current metric depth (D).}
D represents optical-axis depth $d$ in meters, not distance to the
camera center. For positive depth, the fixed log-inverse mapping is
\begin{equation}
q_D(d;d_{\min},d_{\max})=
\frac{\log d_{\max}-\log\!\left(\operatorname{clip}(d,d_{\min},d_{\max})\right)}
{\log d_{\max}-\log d_{\min}},
\qquad d_{\max}=3.5\ \mathrm{m}.
\label{eq:appendix-depth-encoding}
\end{equation}
{The scalar is copied into all three channels without
per-image normalization. We use $d_{\min}=0.04$ m for LIBERO and
$d_{\min}=0.01$ m for all three RoboTwin~2.0 cameras. Nonpositive
RoboTwin depths are encoded as black $(0,0,0)$.}

\Needspace{7\baselineskip}
\paragraph{{Depth backprojection.}}
{To construct normals (N) and pointmaps (P), we first
backproject LIBERO depth maps into 3D points in camera coordinates.} With
intrinsics $K$ and homogeneous pixel $\tilde{\mathbf p}$ in the
OpenCV camera convention, $\mathbf p_C=dK^{-1}\tilde{\mathbf p}$;
the ray is not unit-normalized. The saved images' orientation and
renderer sampling positions are accounted for before backprojection:
a stored pixel $(u_s,v_s)$ maps to
$\tilde{\mathbf p}=(256-0.625-u_s,\ v_s+0.875,\ 1)^{\mathsf T}$.
The camera axes are $+x$ right, $+y$ down, and $+z$ forward.
Encoded targets return to the stored RGB pixel layout.

\paragraph{Camera-frame surface normals (N).}
{Using the backprojected points $\mathbf p_C$, we compute
unit normals in the camera frame.} At each
pixel, horizontal and vertical central differences of $\mathbf p_C$
give a cross product that is normalized and oriented toward the
camera, $\mathbf n^{\mathsf T}(-\mathbf p_C)\geq0$. The three image
channels contain $\operatorname{round}[255(\mathbf n+1)/2]$.
{We exclude neighborhoods with invalid depth or degenerate
local geometry. Valid neighborhoods require center-to-neighbor depth
differences below $0.05\,\mathrm{m}$ and pairwise normal angles of at most
$30^\circ$ among the four center-neighbor triangles. Borders and rejected
neighborhoods are encoded as black.}

\Needspace{10\baselineskip}
\paragraph{Robot-frame pointmaps (P).}
P encodes the robot-base coordinates of visible surface points.
Camera-frame points $\mathbf p_C$ are transformed using camera
extrinsics and the current robot-base pose; X, Y, and Z
occupy the R, G, and B channels. Their fixed central intervals
are $[-0.25,1.00]$, $[-0.45,0.50]$, and $[-0.10,0.50]$ m.
For one coordinate $x$ with interval $[l,h]$, let
$\alpha=224/(h-l)$ and encode
\begin{equation}
c(x)=
\begin{cases}
1+\dfrac{15}{1+\alpha(l-x)/15}, & x<l,\\[4pt]
16+\alpha(x-l), & l\leq x\leq h,\\[4pt]
255-\dfrac{15}{1+\alpha(x-h)/15}, & x>h.
\end{cases}
\label{eq:appendix-pointmap-encoding}
\end{equation}
{Values are clipped to $[1,255]$, rounded to the nearest
integer, and converted to image channels.} The central
interval maps linearly to codes $[16,240]$; smooth tails preserve
out-of-range surface points. The intervals were calibrated on
training trajectories and remain fixed across frames and cameras.
Pixels without a valid 3D point are encoded as black.

\begin{figure}[!htbp]
\centering
\includegraphics[width=\linewidth]{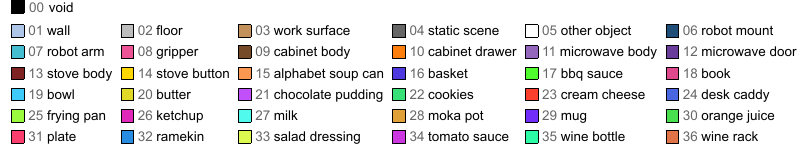}
\caption{The fixed 37-class LIBERO semantic-part ontology and RGB
palette. Colors are shared across all four suites and both cameras.}
\label{fig:appendix-semantic-ontology}
\end{figure}

\paragraph{Semantic parts (S).}
Simulator geometry IDs are mapped through an episode-specific lookup
to a fixed 37-class semantic-part ontology and RGB palette shared
by the four LIBERO suites and both cameras. Classes include
void, static scene surfaces, robot parts, movable objects, and articulated parts
such as drawers, doors, and buttons
(Figure~\ref{fig:appendix-semantic-ontology}). Same-class instances
share a color. S is task independent and covers the whole scene;
it is not a per-scene instance coloring.

\FloatBarrier

\paragraph{Task-conditioned grounding (G).}
G encodes three independent binary roles in R/G/B: the operation
object, goal or anchor, and active articulated part. A pixel can
occupy more than one role. {Roles are assigned to
task-relevant instances using task predicates and the demonstration's
execution stage.} The active stage follows the order in which predicates become
persistently satisfied in that demonstration. The prerequisite
drawer-opening stage requires three consecutive satisfied frames
to be marked complete. Grounding is defined only before full task
completion and only for a camera with at least 64 visible role
pixels.

\paragraph{Future RGB (F) and future depth (FD).}
For current timestep $j$ in a filtered demonstration with
$L$ retained timesteps, the target index is $j_f=\min(j+16,L-1)$. F is the two-view
RGB image at $j_f$; FD is the depth image from the same future state,
encoded and validated exactly as LIBERO D. {Both use the
\emph{current} RGB canvas as their reference.} {When $j+16$
exceeds the final retained timestep, we use the final frame.} The offset counts retained
action steps after no-op filtering, not necessarily 16 original
simulator frames. FD is a single future-depth task; D+F combines
current depth with future RGB.

\subsection{{Label Sources}}
\label{app:visual-data}

\paragraph{LIBERO.}
{To supplement the released LIBERO
demonstrations~\citep{liu2023libero} with visual supervision, we restore
their recorded states in MuJoCo~\citep{todorov2012mujoco} and construct
labels for the agent and wrist views. D comes from simulator-rendered
depth. N is computed from depth and camera intrinsics; P additionally
uses camera extrinsics and the robot-base pose. S maps simulator geometry
IDs to semantic categories. G uses task predicates, object identities,
and demonstration stages to assign task roles. F and FD use RGB and
depth from later states within the same demonstration. All LIBERO target
configurations use the same 1,808 retained training demonstrations.}

\paragraph{RoboTwin~2.0.}
{Depth supervision comes from the RoboTwin~2.0 RGB-D
demonstrations released with Flex-$\pi$~\citep{yan2026flexpi}, which
provide depth maps paired with RGB observations from all three cameras.
Dataset splits and evaluation protocols are specified in
Appendix~\ref{app:robotwin-protocol}.}

\subsection{Prompts and Perception Loss}
\label{app:perception-prompts}

Table~\ref{tab:appendix-visual-prompts} gives the exact perception
prompts. The placeholder \texttt{\{task\}} is replaced by the
manipulation instruction. D, N, S, and P use fixed text; G, F, and
FD incorporate task language. RoboTwin D uses the same D prompt.

\begin{table}[!htbp]
\centering
\caption{{Perception prompts.}}
\label{tab:appendix-visual-prompts}
\small
\setlength{\tabcolsep}{5pt}
\renewcommand{\arraystretch}{1.15}
\begin{tabularx}{\linewidth}{@{}l>{\raggedright\arraybackslash}X@{}}
\toprule
Target & Prompt \\
\midrule
D & Generate a full-scene metric depth map for the current observation. \\
N & Generate a camera-coordinate surface normal map for the current observation. \\
S & Generate a semantic part segmentation map for the current observation using the fixed LIBERO part palette. \\
P & Generate a robot-base-coordinate XYZ point map for the current observation. \\
G & For the robot task: \texttt{\{task\}}. Generate a current-stage task grounding map: red for the operation object, green for the goal or anchor, and blue for the active part. \\
F & A video recorded from a robot's point of view executing the following instruction: \texttt{\{task\}} \\
FD & For the robot task: \texttt{\{task\}}. Generate a full-scene metric depth map for the future observation 16 action steps ahead. \\
\bottomrule
\end{tabularx}
\end{table}

\begingroup
\paragraph{Noise-level sampling and weighting.}
\label{app:flow-schedule}
For the perception objective in Equation~\ref{eq:visual-objective},
we independently sample $u\sim\mathcal U(0,1)$ for each sample
and set $\tau=5u/(1+4u)$. The loss weights follow
\begin{equation}
w(\tau)\propto\exp[-2(\tau-\tfrac12)^2]-\exp(-\tfrac12),
\label{eq:appendix-visual-weight}
\end{equation}
with numerical normalization to unit mean under this sampling schedule.

\paragraph{Loss averaging.}
For LIBERO, perception loss is averaged over target tokens within each
view, equally over valid views within each sample, and then over the
batch. G excludes views without valid grounding targets; samples with
no valid views receive zero perception loss. Black-coded regions in N
and P remain included. RoboTwin depth supervision averages over the full
three-view canvas, including zero-depth regions, giving the main and
two wrist views weights proportional to their areas ($4:1:1$).
\par
\endgroup

\subsection{{Visual-Target Ablation Protocol}}
\label{app:visual-sampling}
\label{app:visual-selection}

{The experiments in Table~\ref{tab:visual-supervision} use the
visual-target ablation configuration in Table~\ref{tab:appendix-training-recipes}.}

\paragraph{{Task sampling.}}
{Table~\ref{tab:appendix-visual-sampling} summarizes task
selection and loss allocation. In multi-target configurations, tasks are
sampled independently for each observation in the training batch.
Each selected task receives one perception forward pass with loss
coefficient 0.1 in Equation~\ref{eq:joint-objective}; when two tasks
are selected, their weighted losses are summed.}

\begin{table}[!htbp]
\centering
\caption{{Perception-task sampling and loss weights.}}
\label{tab:appendix-visual-sampling}
\small
\setlength{\tabcolsep}{5pt}
\renewcommand{\arraystretch}{1.12}
\begin{tabularx}{\linewidth}{@{}l>{\raggedright\arraybackslash}Xcc@{}}
\toprule
Configuration & {Task selection} & {\shortstack{Perception passes\\per action pass}} & {\shortstack{Total perception\\loss coefficient}} \\
\midrule
None & {No perception task} & 0 & 0 \\
Single target & {Fixed task} & 1 & 0.1 \\
D+S / D+P / D+F & {Uniformly sample one task} & 1 & 0.1 \\
D+N+S+P & {Uniformly sample two distinct tasks} & 2 & 0.2 \\
\bottomrule
\end{tabularx}
\end{table}

{The action-update budget is fixed across configurations,
while perception computation and the total perception-loss coefficient
vary as listed in the table.}

\paragraph{{Additional evaluation results.}}
{Table~\ref{tab:supervision-checkpoints} gives the fixed checkpoints
and suite-level independent evaluations underlying
Table~\ref{tab:visual-supervision}.}

{On LIBERO-Plus, the D--F difference is $-0.41$ percentage points
(95\% confidence interval: $[-1.97,1.19]$;
Appendix~\ref{app:task-uncertainty}).}

\begin{table}[!htbp]

\centering
\caption{Fixed checkpoints and independent LIBERO suite results for visual-target ablations (\%).}
\label{tab:supervision-checkpoints}
\fontsize{8}{9.5}\selectfont
\setlength{\tabcolsep}{3pt}
\renewcommand{\arraystretch}{1.08}
\begin{tabular*}{\linewidth}{@{\extracolsep{\fill}}lrrrrr@{}}
\toprule
Target & Step & Spatial & Object & Goal & Long \\
\midrule
None & 32,000 & $95.33\pm0.70$ & $99.40\pm0.53$ & $97.67\pm0.42$ & $95.67\pm0.46$ \\
D & 36,750 & $97.60\pm0.35$ & $99.87\pm0.23$ & $97.87\pm0.76$ & $96.13\pm1.40$ \\
N & 30,000 & $97.60\pm0.53$ & $98.00\pm0.69$ & $95.53\pm0.12$ & $94.93\pm1.01$ \\
S & 36,000 & $97.13\pm0.42$ & $99.67\pm0.31$ & $96.33\pm0.76$ & $95.53\pm1.01$ \\
P & 36,750 & $97.60\pm0.20$ & $98.93\pm0.31$ & $98.33\pm0.61$ & $95.20\pm1.64$ \\
G & 32,000 & $97.07\pm0.23$ & $99.60\pm0.35$ & $93.80\pm0.53$ & $96.00\pm0.60$ \\
F & 30,000 & $97.13\pm0.31$ & $99.33\pm0.31$ & $96.60\pm0.72$ & $94.87\pm0.83$ \\
FD & 34,000 & $97.13\pm0.31$ & $99.87\pm0.12$ & $97.87\pm0.50$ & $97.20\pm0.87$ \\
D+S & 34,000 & $97.40\pm0.35$ & $99.53\pm0.12$ & $96.33\pm0.42$ & $95.67\pm1.70$ \\
D+P & 32,000 & $97.60\pm0.60$ & $99.27\pm0.42$ & $97.80\pm0.35$ & $96.47\pm1.03$ \\
D+F & 28,000 & $97.07\pm0.42$ & $99.60\pm0.53$ & $97.27\pm1.14$ & $95.53\pm1.50$ \\
D+N+S+P & 34,000 & $96.53\pm0.31$ & $99.80\pm0.20$ & $98.07\pm0.81$ & $96.67\pm0.64$ \\
\bottomrule
\end{tabular*}
\par\vspace{3pt}
\begin{minipage}{\linewidth}\footnotesize
Mean $\pm$ SD across three evaluation seeds for each fixed policy.
\end{minipage}
\end{table}

\begingroup
\subsection{Perception Supervision under the {LIBERO Policy Configuration}}
\label{app:reported-config-supervision}

{We additionally compare action-only training (None), current-depth
supervision (D), and future RGB supervision (F)} under the
{LIBERO policy configuration}. {The policies share
the architecture, pretrained initialization, and training demonstrations,
following the LIBERO training recipe in Appendix~\ref{app:training-recipes}.}
None omits the perception forward pass; {D and F each add one perception pass
per action update with loss weight 0.1.} This comparison complements the
{visual-target ablations in}
Table~\ref{tab:visual-supervision}.

\begin{table}[!htbp]

\centering
\caption{Visual supervision under the LIBERO policy configuration (\%).}
\label{tab:reported-config-supervision}
\footnotesize
\setlength{\tabcolsep}{3pt}
\renewcommand{\arraystretch}{1.08}
\begin{tabular*}{\linewidth}{@{\extracolsep{\fill}}lrrr@{}}
\toprule
Metric & None & D & F \\
\midrule
Top LIBERO & $98.80\pm0.22$ & $98.87\pm0.25$ & $98.68\pm0.10$ \\
Final LIBERO & $98.28\pm0.16$ & $98.53\pm0.08$ & $98.18\pm0.25$ \\
Top Plus Clean & $79.49\pm0.85$ & $82.33\pm1.27$ & $81.80\pm2.05$ \\
Final Plus Clean & $79.45\pm0.39$ & $82.35\pm0.85$ & $82.15\pm2.15$ \\
Top Plus Official & $74.63\pm4.03$ & $78.67\pm3.97$ & $77.74\pm6.57$ \\
Final Plus Official & $74.32\pm3.48$ & $79.27\pm3.62$ & $78.64\pm5.82$ \\
\bottomrule
\end{tabular*}
\par\vspace{3pt}
\begin{minipage}{\linewidth}\footnotesize
Mean $\pm$ SD across three training runs.
\end{minipage}
\end{table}

{Both D and F improve mean LIBERO-Plus success over None at Top
and Final checkpoints under both instruction protocols. For Top checkpoints
under Clean-Instruction, the D--F difference is $+0.53$ percentage points
(95\% confidence interval: $[-0.43,1.54]$;
Appendix~\ref{app:task-uncertainty}).}
\par
\endgroup

\begingroup

\subsection{Direct Depth-Regression Control}
\label{app:direct-depth-control}

We compare native flow-based depth supervision (Native flow) with direct
depth regression (DPT regression) using the DPT-style head described in
Appendix~\ref{app:auxiliary-depth-head}.
Both follow the LIBERO policy configuration in
Appendix~\ref{app:training-recipes}, with the same depth labels and aligned
geometric augmentation.

Training costs follow the setup in Appendix~\ref{app:training-cost},
using 50 warm-up and 150 timed updates per run.

Table~\ref{tab:direct-depth-control-results} extends Table~\ref{tab:depth-regression-control}
to Final checkpoints and Official instructions.

\begin{table}[!htbp]

\centering
\caption{Native depth supervision and DPT regression (\%).}
\label{tab:direct-depth-control-results}
\footnotesize
\setlength{\tabcolsep}{3pt}
\renewcommand{\arraystretch}{1.08}
\begin{tabular*}{\linewidth}{@{\extracolsep{\fill}}lrr@{}}
\toprule
Metric & Native flow & DPT regression \\
\midrule
Top LIBERO & $98.87\pm0.25$ & $98.85\pm0.10$ \\
Final LIBERO & $98.53\pm0.08$ & $98.67\pm0.24$ \\
Top Plus Clean & $82.33\pm1.27$ & $80.61\pm0.35$ \\
Final Plus Clean & $82.35\pm0.85$ & $80.27\pm0.59$ \\
Top Plus Official & $78.67\pm3.97$ & $76.72\pm2.95$ \\
Final Plus Official & $79.27\pm3.62$ & $76.39\pm1.96$ \\
\bottomrule
\end{tabular*}
\par\vspace{3pt}
\begin{minipage}{\linewidth}\footnotesize
Mean $\pm$ SD across three training runs.
\end{minipage}
\end{table}

\par
\endgroup
\FloatBarrier

\FloatBarrier
\section{Simulation Evaluation and Baseline Details}
\label{app:baseline-sources}

\subsection{{Standard LIBERO}}
\label{app:libero-protocol}

{The repeated-training comparisons use seeds 42, 142, and 242
for None, current depth, future RGB, the flow-based action adapter, and
DPT regression under the LIBERO policy configuration. Other
GenPA training runs retain the original seed 42.}

\paragraph{{Evaluation protocol.}}
{We evaluate each of the 40 tasks for 50 trials, using the
benchmark's released initial states in their provided order and the
simulator's task-success signal. Episodes allow up to 400 action steps for
Spatial, Object, and Goal and 700 for Long, following 30 initial no-op
steps. The policy predicts 16 actions per call and executes the first 12
before replanning, with binary gripper postprocessing and no temporal
action ensemble. Unless otherwise stated, GenPA benchmark success rates
are measured using uncompiled inference.}

\paragraph{{Checkpoint selection.}}
\label{app:libero-checkpoint-selection}
{We apply one selection rule to all our LIBERO-trained policies,
including both action heads and all perception-supervision configurations.
The candidate set consists of ten checkpoints at 20k, 22k, 24k, 26k, 28k,
30k, 32k, 34k, 36k, and 36.75k action updates. We select the checkpoint
with the most successes across the 2,000 standard LIBERO trials, breaking
ties in favor of the earlier checkpoint. The same checkpoint is used for
LIBERO-Plus, without further selection or fine-tuning.}
{We denote the selected checkpoint as \emph{Top}, the default for
reported results, and the fixed 36,750-step checkpoint as \emph{Final}.
Unless marked independent, our standard LIBERO results use the same initial
states as checkpoint selection. Complete checkpoint scans are provided in
Appendix~\ref{app:libero-validation}.}

\paragraph{{Independent LIBERO evaluation.}}
{Since standard LIBERO scores are used for checkpoint selection,
we re-evaluate the Top checkpoints from training seed 42 using three
evaluation seeds (1042, 1142, and 1242). Each seed provides 50 new initial
states per task, yielding 6,000 episodes per policy. The states satisfy the
original task constraints and are disjoint across sets and from those used
for checkpoint selection. All policies use the same three sets.}

\paragraph{{Baseline sources.}}
{Standard LIBERO baseline scores come from published reports;
Table~\ref{tab:baseline-standard-sources} identifies their sources and model
variants. Their training and checkpoint-selection procedures may differ
from ours. Robot PT denotes robot-data pretraining before benchmark
adaptation, excluding generic image, video, or vision-language pretraining.}

\begin{table}[!htbp]
\centering
\caption{Sources and selected configurations for the LIBERO comparison.}
\label{tab:baseline-standard-sources}
\footnotesize
\setlength{\tabcolsep}{4pt}
\renewcommand{\arraystretch}{1.13}
\begin{tabularx}{\linewidth}{@{}>{\raggedright\arraybackslash}p{0.19\linewidth}>{\raggedright\arraybackslash}p{0.31\linewidth}>{\raggedright\arraybackslash}X@{}}
\toprule
Method & Score source & Configuration used \\
\midrule
OpenVLA-OFT & \citet{kim2025openvlaoft}, Table~XIV &
{Joint four-suite policy initialized from robot-pretrained OpenVLA.} \\
$\pi_{0.5}$ & \citet{physicalintelligence2026openpilibero}, official LIBERO release;
model described by \citet{physicalintelligence2025pi05} &
{Released 30k-step flow policy.} \\
VLA-Adapter-Pro & \citet{wang2026vlaadapter}, arXiv v2, Table~5 &
{Pro variant; four suite-specific checkpoints
(\href{https://github.com/OpenHelix-Team/VLA-Adapter/blob/main/README.md}{official release}).} \\
SimVLA & \citet{luo2026simvla}, Table~2 &
{Large action-head configuration with generic VLM initialization.} \\
VLANeXt & \citet{wu2026vlanext}, arXiv v3, Table~2 &
Base LIBERO configuration. Its separate DROID-pretrained real-world variant is not used. \\
Qwen-RobotManip & \citet{qwen2026robotmanip}, Table~3 &
{Base model, not the Context variant; only overall success is reported.} \\
LaWAM & \citet{chen2026lawam}, Table~1 &
Robot-video-pretrained policy with a truncated Qwen3-VL backbone and a latent world model. \\
Cosmos Policy & \citet{kim2026cosmospolicy}, Table~1 &
{Author-reported mean over 6,000 evaluation episodes.} \\
Fast-WAM & \citet{yuan2026fastwam}, arXiv v2, Table~2 &
{Fast-WAM, not its Joint or IDM variant.} \\
ABot-M0.5 & \citet{amap2026abotm05}, Table~5 &
Robot-pretrained model with Wan2.2-5B initialization. \\
ImageWAM-4B & \citet{zhang2026imagewam}, Table~2 &
{FLUX.2 4B configuration.} \\
GenPA & This work &
{Depth-supervised policy trained jointly on all four suites.} \\
\bottomrule
\end{tabularx}
\end{table}

\paragraph{{Parameter counts.}}
\label{app:parameter-counts}
{Unless specified below, counts in Table~\ref{tab:libero-main}
follow the papers or official model releases listed in
Table~\ref{tab:baseline-standard-sources}. Backbone sizes use reported model
sizes; separate text encoders and VAEs are excluded for visual generation
backbones. Head counts include input/output and conditioning projections,
but exclude separately implemented proprioception projectors and backbone
query tokens. GenPA's adapter count follows Appendix~\ref{app:action-heads}.}

{We count OpenVLA-OFT's 151M-parameter head from its
\href{https://github.com/moojink/openvla-oft/blob/main/prismatic/models/action_heads.py}{official implementation},
confirming it against the loaded model. Counts from public weights use
VLA-Adapter-Pro's
\href{https://huggingface.co/VLA-Adapter/LIBERO-Spatial-Pro}{Spatial-Pro checkpoint}
(1.25B VLM, 218M head) and LaWAM's
\href{https://huggingface.co/jialei02/lawam_libero_sft_release}{LIBERO checkpoint}
(1.52B VLM, 306M head). VLANeXt's 612M head is computed from its
\href{https://github.com/DravenALG/VLANeXt/blob/main/config/VLANeXt-B_config.yaml}{official base configuration}
and implementation. SimVLA's approximately 300M head follows the authors'
estimate.}

{VLA-Adapter-Pro's VLM count includes its vision encoders,
language model, and projector; its quoted 0.5B refers only to the language
model. LaWAM's checkpoint additionally contains a 228M latent world decoder
and an 86M DINO encoder, identified separately from its action head;
training-only modules are excluded. Cosmos Policy uses its diffusion
backbone for action generation, so its head entry is \emph{Shared}.
Qwen-RobotManip and ABot-M0.5 have unverified head counts because complete
configurations or accessible weights for the corresponding LIBERO policies
were unavailable at the time of inspection; their entries remain blank.}

\FloatBarrier
\subsection{LIBERO-Plus: Instruction Processing, Re-evaluation, and Published Results}
\label{app:plus-protocol}

\paragraph{Instruction processing.}
{We refer to evaluation with released instructions as \emph{Official}
and with the processing below as \emph{Clean}.}
For 8,493 non-language instances, the released instructions contain appended
perturbation metadata. We remove only the exact perturbation suffix
and preserve the canonical task instruction. All 1,537 Language-category
instructions remain unchanged, including any numbers that are part of the
intended instruction. Table~\ref{tab:plus-instruction-examples} shows an
example from each perturbation category before and after processing.
{Flex-$\pi$ also identifies this instruction
issue~\citep{yan2026flexpi}.}

\begin{table}[!htbp]
\centering
\caption{Representative LIBERO-Plus instructions before and after processing.
The Language instruction is preserved; in the other categories only appended
perturbation identifiers are removed.}
\label{tab:plus-instruction-examples}
\footnotesize
\setlength{\tabcolsep}{3pt}
\renewcommand{\arraystretch}{1.08}
\begin{tabularx}{\linewidth}{@{}p{0.18\linewidth}>{\raggedright\arraybackslash}X>{\raggedright\arraybackslash}X@{}}
\toprule
Perturbation & Released instruction & Clean-Instruction \\
\midrule
Camera & put both the alphabet soup and the tomato sauce in the basket view
0 0 100 2 10 initstate 0 & put both the alphabet soup and the tomato sauce
in the basket \\
Robot initialization & put both the alphabet soup and the tomato sauce in
the basket view 0 0 100 0 0 initstate 51 & put both the alphabet soup and
the tomato sauce in the basket \\
Language & before we start cooking dont forget to put both the alphabet soup
and the tomato sauce in the basket & before we start cooking dont forget to
put both the alphabet soup and the tomato sauce in the basket \\
Lighting & turn on the stove and put the moka pot on it light 19 & turn on
the stove and put the moka pot on it \\
Background & turn on the stove and put the moka pot on it table 19 & turn on
the stove and put the moka pot on it \\
Sensor noise & put both the alphabet soup and the tomato sauce in the basket
view 0 0 100 0 0 initstate 0 noise 5 & put both the alphabet soup and the
tomato sauce in the basket \\
Object layout & turn on the stove and put the moka pot on it add 7 & turn on
the stove and put the moka pot on it \\
\bottomrule
\end{tabularx}
\end{table}

\paragraph{{Evaluation protocol.}}
{{Across our LIBERO-Plus evaluations,} we retain the released environments and run one
trial per task instance for each policy, using environment seed 42 and the
first released initial state.}
Rendering uses two $256\times256$ views, followed by each policy's native
preprocessing. We use 30 initial no-op steps and limits of 400 action steps
for Spatial, Object, and Goal and 700 for Long. The seven categories contain
1,599 Camera, 1,550 Robot, 1,537 Language, 1,142 Light, 1,076 Background,
1,601 Noise, and 1,525 Layout instances. Overall success divides total
successes by 10,030; category scores use their respective instance counts.
{All evaluations cover all 10,030 instances.}
{Unless otherwise stated,} success rates use uncompiled inference.
The external policies use released LIBERO checkpoints; we do not fine-tune
them on LIBERO-Plus.

\begin{table}[!htbp]
\centering
\caption{{Checkpoints and inference settings for the LIBERO-Plus comparison in Table~\ref{tab:robust-main}.}
$H$ is the predicted action horizon; $R$ is the number of executed actions before replanning;
$K$ is the number of action-denoising steps.}
\label{tab:baseline-plus-sources}
\footnotesize
\setlength{\tabcolsep}{3pt}
\renewcommand{\arraystretch}{1.12}
\begin{tabularx}{\linewidth}{@{}>{\raggedright\arraybackslash}p{0.18\linewidth}>{\raggedright\arraybackslash}Xrrr@{}}
\toprule
Policy & Checkpoint and inference path & $H$ & $R$ & $K$\\
\midrule
OpenVLA-OFT & Official joint four-suite 300k checkpoint; regression~\citep{kim2025openvlaoft}. & 8 & 8 & ---\\
$\pi_{0.5}$ & Official \texttt{pi05\_libero}; released PyTorch conversion and native transforms~\citep{physicalintelligence2025pi05}. & 10 & 5 & 10\\
VLANeXt & Released suite-specific checkpoints; official LIBERO-Plus entry point~\citep{wu2026vlanext}. & 8 & 8 & 5\\
Cosmos Policy & Released 2B LIBERO policy, retaining parallel future/value generation~\citep{kim2026cosmospolicy}. & 16 & 16 & 5\\
Fast-WAM & Released two-view 224-pixel LIBERO checkpoint; action-only inference~\citep{yuan2026fastwam}. & 32 & 10 & 10\\
ImageWAM & Released FLUX.2 4B LIBERO checkpoint; default Prefix-Only inference~\citep{zhang2026imagewam}. & 16 & 12 & 10\\
GenPA & Depth-supervised 32k checkpoint; direct regression. & 16 & 12 & ---\\
\bottomrule
\end{tabularx}
\end{table}

{We evaluate all policies on the same environment instances
with the same processed task instructions. Policies differ in pretraining and retain their native image
transforms, action normalization, prompt templates, action horizons, and
replanning frequencies.} Fast-WAM uses a noise-schedule
shift of 5.0. Cosmos Policy retains its native parallel future/value outputs
without additional autoregressive planning or candidate search.
{VLANeXt uses five denoising steps, following its public
LIBERO-Plus entry point rather than the checkpoint default of ten.
ImageWAM's released LIBERO evaluator uses Prefix-Only inference by default,
which we follow for re-evaluation. This path omits the image-denoising step
used in its paper's main results.}

At the time of evaluation, Qwen-RobotManip did not provide a public model
checkpoint, and we could not obtain an accessible ABot-M0.5 checkpoint
fine-tuned for LIBERO. We exclude these two methods from the common-protocol
re-evaluation while retaining their published results in the standard
LIBERO comparison and Table~\ref{tab:robust-published}.

\paragraph{Published LIBERO-Plus scores.}
{Table~\ref{tab:robust-published} lists published baseline
scores alongside GenPA's evaluation with the released instructions.
These results do not constitute a common-protocol comparison: instruction
processing and aggregation are not uniformly documented across sources.
Table~\ref{tab:baseline-plus-published-sources} records each source,
configuration, and aggregation method. Differences from our
Clean-Instruction re-evaluations cannot be attributed solely to instruction
cleaning, since checkpoints, inference paths, and evaluation settings may
also differ.}

\noindent\begin{minipage}{\linewidth}
\centering
\makeatletter\def\@captype{table}\makeatother
\caption{Published LIBERO-Plus baseline scores and GenPA evaluation with released instructions (\%).}
\label{tab:robust-published}
\vspace{3pt}
\fontsize{8}{9.2}\selectfont
\setlength{\tabcolsep}{2.3pt}
\renewcommand{\arraystretch}{1.05}
\begin{tabular*}{\linewidth}{@{\extracolsep{\fill}}lcrrrrrrrr@{}}
\toprule
Method & \shortstack{Robot\\PT} & Camera & \shortstack{Robot\\Init.} & Lang. & Light & Backgr. & Noise & Layout & Overall\\
\midrule
OpenVLA-OFT$^{j}$ & \robotpt & 55.6 & 21.7 & 81.0 & 92.7 & 91.0 & 78.6 & 68.7 & 67.9\\
$\pi_{0.5}$ & \robotpt & 75.4 & 77.5 & 85.6 & 96.9 & 94.6 & 89.7 & 85.7 & 85.7\\
VLANeXt & \norobotpt & 90.1 & 65.2 & 81.0 & 96.0 & 82.1 & 93.6 & 80.1 & 83.7\\
Qwen-RobotManip & \robotpt & 87.2 & 75.5 & 85.6 & 96.6 & 97.7 & 97.7 & 87.3 & 89.0\\
Cosmos Policy & \norobotpt & 75.8 & 63.3 & 81.7 & 96.5 & 88.9 & 92.7 & 82.2 & 82.2\\
Fast-WAM & \norobotpt & 16.4 & 44.5 & 68.9 & 78.2 & 53.7 & 37.7 & 60.7 & 51.5\\
ABot-M0.5$^{\dagger}$ & \robotpt & 70.5 & 87.4 & 88.6 & 94.0 & 89.7 & 75.5 & 85.2 & 83.4\\
ImageWAM-4B$^{\dagger}$ & \norobotpt & 80.8 & 50.3 & 91.4 & 98.1 & 85.5 & 93.8 & 80.5 & 83.1\\
\textbf{GenPA} & \norobotpt & 81.18 & 67.29 & 84.97 & 97.64 & 80.11 & 91.69 & 80.92 & 83.01\\
\bottomrule
\end{tabular*}
\vspace{2pt}
\begin{minipage}{\linewidth}\scriptsize
{Robot PT is defined in Table~\ref{tab:libero-main}; $j$ denotes joint four-suite training.
$\dagger$: The published overall aggregation cannot be reconstructed from the reported breakdowns.}
\end{minipage}
\end{minipage}

\begin{table}[!htbp]
\centering
\caption{Sources and aggregation of the published LIBERO-Plus scores in
Table~\ref{tab:robust-published}.}
\label{tab:baseline-plus-published-sources}
\footnotesize
\setlength{\tabcolsep}{4pt}
\renewcommand{\arraystretch}{1.13}
\begin{tabularx}{\linewidth}{@{}>{\raggedright\arraybackslash}p{0.19\linewidth}>{\raggedright\arraybackslash}p{0.31\linewidth}>{\raggedright\arraybackslash}X@{}}
\toprule
Method & Evaluation source & Selection and aggregation \\
\midrule
OpenVLA-OFT & \citet{fei2026liberoplus}, Table~2 and official leaderboard &
{Joint four-suite policy (OpenVLA-OFT\_m).} \\
$\pi_{0.5}$ & \citet{zhang2026wamrobustness}, Table~4 &
{Third-party evaluation.} \\
VLANeXt & \citet{wu2026vlanext}, Table~3 &
{Approximate overall score, reweighted from rounded suite scores using released instance counts.} \\
Qwen-RobotManip & \citet{qwen2026robotmanip}, Table~4 &
{Base configuration.} \\
Cosmos Policy & \citet{zhang2026wamrobustness}, Table~4 &
{Third-party evaluation.} \\
Fast-WAM & \citet{zhang2026wamrobustness}, Table~4 &
{Third-party evaluation; overall is the mean across four LIBERO suites.} \\
ABot-M0.5 & \citet{amap2026abotm05}, Table~6 &
The reported overall weighting is not established as instance-pooled. \\
ImageWAM-4B & \citet{zhang2026imagewam}, Table~3 &
The 4B result's overall weighting is not established as instance-pooled. \\
GenPA & This work &
Released instructions; 8,326 successes over 10,030 instances from the same 32k-step checkpoint. \\
\bottomrule
\end{tabularx}
\end{table}

\paragraph{{Uncertainty in LIBERO-Plus comparisons.}}
\label{app:task-uncertainty}
{For the internal regression--flow and depth--future-RGB comparisons
on LIBERO-Plus, we report pointwise 95\%
percentile confidence intervals from 20,000 paired task-cluster bootstrap
replicates. Within each of the four suites, we resample its ten original
tasks with replacement, retaining all perturbation instances and paired
policy outcomes and preserving instance-pooled scoring. For three-run
comparisons, we average the three paired differences within each replicate.
These intervals quantify task-level uncertainty for fixed policies;
training-run variability is reported separately.}

\FloatBarrier
\begingroup

\subsection{Checkpoint Sensitivity and Independent LIBERO Results}
\label{app:libero-validation}

Table~\ref{tab:multiseed-scan-visual} gives complete checkpoint scans
for the supervision comparison in Appendix~\ref{app:reported-config-supervision}.
Table~\ref{tab:multiseed-scan-heads} covers Flow
(Appendix~\ref{app:action-readout-results}) and DPT
(Appendix~\ref{app:direct-depth-control}), with their shared D baseline in
Table~\ref{tab:multiseed-scan-visual}.

\begin{table}[!htbp]

\centering
\caption{Standard-LIBERO checkpoint scans for None, D, and F (\%). Column suffixes denote training seeds.}
\label{tab:multiseed-scan-visual}
\fontsize{8}{9.5}\selectfont
\setlength{\tabcolsep}{3pt}
\renewcommand{\arraystretch}{1.08}
\begin{tabular*}{\linewidth}{@{\extracolsep{\fill}}lrrrrrrrrr@{}}
\toprule
Step & None 42 & None 142 & None 242 & D 42 & D 142 & D 242 & F 42 & F 142 & F 242 \\
\midrule
20,000 & 98.25 & 97.60 & 97.80 & 97.60 & 98.30 & 98.30 & 98.25 & 97.30 & 98.50 \\
22,000 & 97.90 & 97.30 & 97.00 & 97.20 & 97.95 & 97.85 & 97.85 & 98.05 & 97.10 \\
24,000 & 97.65 & 97.40 & 98.15 & 98.45 & \textbf{98.90} & 98.45 & 97.15 & 98.55 & 97.90 \\
26,000 & 98.25 & 98.10 & 98.30 & 98.65 & 98.65 & 98.20 & 98.25 & 98.00 & \textbf{98.60} \\
28,000 & \textbf{98.95} & 98.30 & \textbf{98.55} & 99.00 & 98.05 & \textbf{98.60} & 98.65 & 98.10 & 98.60 \\
30,000 & 98.10 & 98.35 & 97.80 & 98.65 & 98.40 & 98.40 & 98.10 & 98.30 & 97.70 \\
32,000 & 98.95 & 98.30 & 98.55 & \textbf{99.10} & 98.25 & 98.20 & 98.75 & \textbf{98.65} & 98.30 \\
34,000 & 98.75 & \textbf{98.90} & 98.40 & 98.75 & 98.30 & 98.50 & 98.20 & 98.60 & 97.90 \\
36,000 & 98.50 & 98.60 & 98.55 & 98.60 & 98.55 & 98.40 & \textbf{98.80} & 98.55 & 98.25 \\
36,750 & 98.40 & 98.10 & 98.35 & 98.45 & 98.60 & 98.55 & 98.35 & 98.30 & 97.90 \\
\bottomrule
\end{tabular*}
\par\vspace{3pt}
\begin{minipage}{\linewidth}\footnotesize
Bold marks the selected checkpoint.
\end{minipage}
\end{table}

\begin{table}[!htbp]

\centering
\caption{Standard-LIBERO checkpoint scans for Flow and DPT (\%). Column suffixes denote training seeds.}
\label{tab:multiseed-scan-heads}
\fontsize{8}{9.5}\selectfont
\setlength{\tabcolsep}{3pt}
\renewcommand{\arraystretch}{1.08}
\begin{tabular*}{\linewidth}{@{\extracolsep{\fill}}lrrrrrr@{}}
\toprule
Step & Flow 42 & Flow 142 & Flow 242 & DPT 42 & DPT 142 & DPT 242 \\
\midrule
20,000 & 93.15 & 94.75 & 83.60 & 97.65 & 97.00 & 97.80 \\
22,000 & 94.20 & 93.40 & 93.20 & \textbf{98.95} & 97.90 & 97.75 \\
24,000 & 95.35 & 93.70 & 93.95 & 97.55 & 98.15 & 96.85 \\
26,000 & 94.90 & 95.55 & 96.45 & 98.10 & 98.00 & 98.35 \\
28,000 & 96.95 & 97.25 & 97.45 & 98.25 & 97.90 & 98.55 \\
30,000 & 97.85 & 98.00 & 97.25 & 98.30 & \textbf{98.75} & 98.45 \\
32,000 & 97.80 & \textbf{98.20} & \textbf{98.20} & 98.95 & 98.30 & 98.40 \\
34,000 & 97.95 & 97.90 & 97.90 & 98.90 & 98.25 & 98.75 \\
36,000 & 97.50 & 97.90 & 98.00 & 98.15 & 98.60 & 98.25 \\
36,750 & \textbf{98.40} & 97.60 & 98.00 & 98.75 & 98.40 & \textbf{98.85} \\
\bottomrule
\end{tabular*}
\par\vspace{3pt}
\begin{minipage}{\linewidth}\footnotesize
Bold marks the selected checkpoint. Flow denotes the flow-based action head.
\end{minipage}
\end{table}

\FloatBarrier

Table~\ref{tab:independent-main} reports independent LIBERO results for
the selected checkpoints from training seed 42, following the protocol in
Appendix~\ref{app:libero-protocol}.

\begin{table}[!htbp]

\centering
\caption{Independent evaluation under the LIBERO policy configuration (\%).}
\label{tab:independent-main}
\fontsize{8}{9.5}\selectfont
\setlength{\tabcolsep}{3pt}
\renewcommand{\arraystretch}{1.08}
\begin{tabular*}{\linewidth}{@{\extracolsep{\fill}}lrrrrrr@{}}
\toprule
Policy & Step & Spatial & Object & Goal & Long & Overall \\
\midrule
None & 28,000 & $96.40\pm0.35$ & $99.67\pm0.31$ & $97.60\pm0.53$ & $96.87\pm0.42$ & $97.63\pm0.24$ \\
D (regression action head) & 32,000 & $97.60\pm0.20$ & $99.73\pm0.31$ & $98.20\pm0.40$ & $97.80\pm1.11$ & $98.33\pm0.29$ \\
F & 36,000 & $97.13\pm0.64$ & $99.27\pm0.12$ & $97.73\pm0.70$ & $97.07\pm0.12$ & $97.80\pm0.05$ \\
D (flow action head) & 36,750 & $96.60\pm0.35$ & $99.00\pm0.35$ & $97.07\pm0.90$ & $95.47\pm0.31$ & $97.03\pm0.33$ \\
DPT regression & 22,000 & $97.47\pm0.31$ & $99.73\pm0.12$ & $98.47\pm0.83$ & $96.80\pm0.40$ & $98.12\pm0.33$ \\
\bottomrule
\end{tabular*}
\par\vspace{3pt}
\begin{minipage}{\linewidth}\footnotesize
Mean $\pm$ SD across three evaluation seeds for each fixed policy.
\end{minipage}
\end{table}

\FloatBarrier
\endgroup

\FloatBarrier
\subsection{RoboTwin 2.0: Evaluation, Baselines, and Adapter Width}
\label{app:robotwin-protocol}
\paragraph{{Training data and evaluation.}}
{Table~\ref{tab:robotwin-main} compares policies trained
jointly across 50 tasks using both clean and randomized demonstrations.}
We use the RGB-D-annotated RoboTwin
dataset released with Flex-$\pi$~\citep{yan2026flexpi}, containing 2,500
clean and 25,000 randomized demonstrations.
{We randomly split the demonstrations at the episode level
with seed 42 into 27,225 training episodes and 275 validation episodes,
without stratification by task or scene condition.}
{We evaluate 100 trials per task under each of the Clean
and Randomized conditions.}

\paragraph{Published baseline results.}
{The $\pi_{0.5}$ and Fast-WAM scores are from
\citet{yuan2026fastwam}, arXiv v2, Table~1; the latter is the
action-only Fast-WAM variant. ImageWAM scores are from
\citet{zhang2026imagewam}, Table~1, using the FLUX.2 4B
configuration described in Section~4.1 of that paper.}
{Qwen-RobotManip uses the Base configuration from
\citet{qwen2026robotmanip}, arXiv v2, Table~3, where Easy/Hard correspond to
Clean/Randomized.}
{Motus scores are from \citet{bi2025motus}, Table~2,
for the robot-pretrained policy fine-tuned on 2,500 clean and 25,000
randomized demonstrations.}
{Clean and Randomized scores are taken from published
reports. Where a source does not report the Avg. at the displayed
precision, we compute the arithmetic mean of its two condition scores
and round to the displayed precision. These are not matched
re-evaluations.}

\paragraph{Action-adapter width.}
{The RoboTwin results in Table~\ref{tab:robotwin-main}
use an action adapter with a hidden width of 1,024, compared with 512
for the LIBERO policy. To assess the effect of this increase, we compare
both widths on RoboTwin in Table~\ref{tab:robotwin-head-width}.
Both adapters follow the
design in Appendix~\ref{app:action-heads} and use the same backbone,
data split, depth supervision, and training recipe, with training seed 42.}
{For each width, we save and screen 12 checkpoints at
28k, 32k, 36k, 42k, 46k, 50k, 54k, 56k, 60k, 64k, 68k, and 70,847
updates. We select the checkpoint with the highest average success across
Clean and Randomized conditions, using 25 trials per task per condition,
then evaluate the selected checkpoint with the 100-trial protocol above.}

\begin{table}[htbp]
\centering
\caption{{RoboTwin action-adapter width comparison (success rates in \%).
The 1,024-width configuration is used in Table~\ref{tab:robotwin-main}.}}
\label{tab:robotwin-head-width}
\small
\setlength{\tabcolsep}{6pt}
\begin{tabular}{lrrrrr}
\toprule
Adapter width & Checkpoint & Parameters & Clean & Rand. & Avg.\\
\midrule
512 & 64k & 18.69M & 91.04 & 89.84 & 90.44\\
1,024 & 64k & 71.45M & 92.08 & 90.82 & 91.45\\
\bottomrule
\end{tabular}
\end{table}

\FloatBarrier
\section{{Action Readout and Inference Efficiency}}
\label{app:efficiency}

{This appendix supplements Section~\ref{sec:efficiency}
with {action-readout comparisons,} timing protocols,
{latency and memory measurements}, and compiled-policy evaluation.} {The timed policies follow the LIBERO inference
settings in Table~\ref{tab:baseline-plus-sources}.}

\begingroup

\subsection{Action-Readout Comparison}
\label{app:action-readout-results}

We compare the regression and flow-based adapters in
Appendix~\ref{app:action-heads}, retaining current-depth supervision and the
LIBERO policy configuration in Appendix~\ref{app:training-recipes}.
Table~\ref{tab:action-readout-results} extends Table~\ref{tab:e9-action-readout} to Final
checkpoints and Official instructions. Regression has higher mean standard
LIBERO success, whereas the Plus ranking changes with checkpoint and
instruction protocol. For Top checkpoints under Clean-Instruction, regression exceeds flow by
$0.84$ percentage points on LIBERO-Plus (95\% confidence interval:
$[-0.49,2.24]$; Appendix~\ref{app:task-uncertainty}).

\begin{table}[!htbp]

\centering
\caption{Action-readout success across checkpoint and instruction settings (\%).}
\label{tab:action-readout-results}
\footnotesize
\setlength{\tabcolsep}{3pt}
\renewcommand{\arraystretch}{1.08}
\begin{tabular*}{\linewidth}{@{\extracolsep{\fill}}lrr@{}}
\toprule
Metric & Regression & Flow \\
\midrule
Top LIBERO & $98.87\pm0.25$ & $98.27\pm0.12$ \\
Final LIBERO & $98.53\pm0.08$ & $98.00\pm0.40$ \\
Top Plus Clean & $82.33\pm1.27$ & $81.49\pm1.18$ \\
Final Plus Clean & $82.35\pm0.85$ & $82.46\pm0.43$ \\
Top Plus Official & $78.67\pm3.97$ & $79.47\pm2.42$ \\
Final Plus Official & $79.27\pm3.62$ & $79.98\pm2.60$ \\
\bottomrule
\end{tabular*}
\par\vspace{3pt}
\begin{minipage}{\linewidth}\footnotesize
Mean $\pm$ SD across three training runs.
\end{minipage}
\end{table}

\FloatBarrier
\endgroup

\subsection{Measurement Protocol}
\label{app:timing-protocol}

We measure batch-one policy calls on an NVIDIA H100 80GB HBM3 (700\,W power
limit) and an NVIDIA L40S (350\,W power limit). Both systems use PyTorch
2.7.1 and CUDA 11.8. {Models use BF16 for their main
computations while retaining implementation-specific higher-precision
operations. Each configuration is benchmarked in three independent runs,
each in a fresh process with exclusive GPU access and four CPU inference
threads.} After 50 warmup calls, each
process measures the same 200 observations from 40 LIBERO tasks (five
observations per task). We report the mean and the sample standard deviation
of the three process means. Latencies are per native action chunk, whose
horizon differs across policies. {\emph{Uncompiled} denotes
PyTorch eager execution; \emph{compiled} denotes execution with
\texttt{torch.compile} and TorchInductor, with model-specific compilation
scope described below.}

For the main comparison, synchronized wall-clock timing begins with
decoded CPU RGB images, raw proprioception, and either a raw instruction
{or cached text features, as specified below.}
It ends with the CPU action
chunk after native postprocessing. Image preprocessing, CPU--GPU transfers,
model execution, and action denormalization are included. Image-file
decoding, model loading, compilation, and environment or robot communication
are excluded. Only observation-independent text features may be reused
across calls; visual features and observation-conditioned representations
are recomputed for every observation. These offline measurements do not
measure robot request--response time or physical control frequency.

The total labels in Figure~\ref{fig:latency-architecture} and the L40S
totals below come from complete calls without internal instrumentation.
The colored component intervals in that figure come from a separate
profiling pass with the same inputs and warmup procedure. They are not
rescaled to match the independently timed totals.

\subsection{{Component Profiling and Compilation}}
\label{app:timing-components}

{Table~\ref{tab:latency-stage-boundaries} maps each policy's measured stages
to the \emph{Vision}, \emph{Backbone}, and \emph{Action} labels in
Figure~\ref{fig:latency-architecture}. GenPA, Fast-WAM, and ImageWAM reuse
independently encoded text features across observations; OpenVLA-OFT and
$\pi_{0.5}$ do not.}

\begin{table}[!htbp]
\centering
\caption{Model-specific component boundaries for Figure~\ref{fig:latency-architecture}.}
\label{tab:latency-stage-boundaries}
\footnotesize
\setlength{\tabcolsep}{3pt}
\renewcommand{\arraystretch}{1.12}
\begingroup
\begin{tabularx}{\linewidth}{@{}l *{3}{>{\raggedright\arraybackslash}X}@{}}
\toprule
Policy & Vision & Backbone & Action\\
\midrule
OpenVLA-OFT & Vision encoder and projector & Multimodal Llama forward, including parallel action slots &
Proprioception projector and regression head\\
\addlinespace[3pt]
$\pi_{0.5}$ & SigLIP encoding and visual projection & VLM prefix forward &
Ten-step action expert\\
\addlinespace[3pt]
Fast-WAM & VAE encoding and visual projection & Video input preparation and backbone prefill &
Ten-step action loop\\
\addlinespace[3pt]
ImageWAM & VAE encoding and visual projection & FLUX Prefix-Only prefill &
Ten-step action loop\\
\addlinespace[3pt]
GenPA & VAE encoding and visual projection & One FLUX forward &
Complete regression adapter\\
\bottomrule
\end{tabularx}
\endgroup
\end{table}

\paragraph{{Profiling boundaries.}}
{Component latencies are measured with nonoverlapping CUDA-event
intervals within full policy calls rather than by running each module in
isolation. \emph{Other} is the profiled complete-call latency minus the
sum of the timed components. It includes remaining preprocessing,
transfers, and postprocessing.}

\paragraph{{Model-specific boundaries.}}
The official PyTorch \(\pi_{0.5}\) configuration encodes two observed images
and a masked zero-image slot. {Its \emph{Action} interval
includes ten expert calls; Euler updates and loop control fall in
\emph{Other} without compilation, whereas compiled \emph{Action} also
includes Euler updates. Compiled \emph{Vision} and \emph{Backbone} are timed
jointly because they share a captured graph.}

\paragraph{{Compilation.}}
{GenPA and ImageWAM compile VAE encoding, backbone conditioning,
and action decoding, including the complete eight-step loop for GenPA's
flow variant.
OpenVLA-OFT compiles its vision encoder, projectors, Llama backbone, and
regression head. These policies use the \texttt{default} compilation mode.
Fast-WAM uses \texttt{reduce-overhead} for backbone prefill and individual
action steps, leaving its VAE uncompiled. $\pi_{0.5}$ uses
\texttt{max-autotune} for its full action-inference procedure.
Native preprocessing and attention implementations are
retained. These cross-policy comparisons characterize the measured
implementations rather than isolate the causal effect of a single module.}
\FloatBarrier

\subsection{H100 Component and Total Latency}
\label{app:timing-h100}

Table~\ref{tab:latency-h100-breakdown} gives the component
and independently timed total latencies underlying Figure~\ref{fig:latency-architecture}.
\begin{table}[!htbp]
\centering
\caption{H100 latency breakdown and independently timed totals (ms per native action chunk).}
\label{tab:latency-h100-breakdown}
\footnotesize
\setlength{\tabcolsep}{3pt}
\renewcommand{\arraystretch}{1.12}
\begin{tabular*}{\linewidth}{@{\extracolsep{\fill}}llrrrrr@{}}
\toprule
Policy & Execution & Vision & Backbone & Action & Other & Total\\
\midrule
OpenVLA-OFT & Uncompiled & 27.57 & 26.75 & 1.05 & 16.97 & $72.12 \pm 1.29$\\
 & Compiled & 14.91 & 21.64 & 0.32 & 16.70 & $53.36 \pm 0.29$\\
\addlinespace[3pt]
$\pi_{0.5}$ & Uncompiled & 24.61 & 13.58 & 121.09 & 7.76 & $166.10 \pm 0.96$\\
 & Compiled & \multicolumn{2}{c}{14.09 (joint)} & 14.20 & 8.75 & $36.59 \pm 0.06$\\
\addlinespace[3pt]
Fast-WAM & Uncompiled & 5.05 & 32.34 & 271.42 & 11.81 & $322.07 \pm 9.53$\\
 & Compiled & 4.96 & 10.22 & 45.25 & 11.08 & $71.34 \pm 0.23$\\
\addlinespace[3pt]
ImageWAM & Uncompiled & 4.82 & 24.30 & 121.49 & 10.30 & $159.51 \pm 1.69$\\
 & Compiled & 2.37 & 15.78 & 28.73 & 10.20 & $57.39 \pm 1.27$\\
\addlinespace[3pt]
GenPA & Uncompiled & 4.71 & 16.04 & 0.62 & 6.62 & $27.86 \pm 0.32$\\
 & Compiled & 2.29 & 10.46 & 0.46 & 7.83 & $20.94 \pm 0.47$\\
\bottomrule
\end{tabular*}
\par\vspace{3pt}
\begin{minipage}{\linewidth}\scriptsize
Components come from separate profiling passes; Total reports independently timed complete calls
(mean $\pm$ SD across three process means), so components need not sum to Total.
Vision and Backbone are timed jointly for compiled $\pi_{0.5}$.
\end{minipage}
\end{table}

\FloatBarrier

\subsection{Complete Policy-Call Latency on L40S}
\label{app:timing-l40s}

Table~\ref{tab:latency-l40s} reports the L40S complete-call results for the
five policies in Figure~\ref{fig:latency-architecture} and the GenPA flow
readout in Table~\ref{tab:e9-action-readout}.

\begin{table}[!htbp]
\centering
\caption{{L40S action-chunk latency (ms; mean $\pm$ SD).}}
\label{tab:latency-l40s}
\footnotesize
\setlength{\tabcolsep}{3pt}
\renewcommand{\arraystretch}{1.08}
\begin{tabular*}{\linewidth}{@{\extracolsep{\fill}}lrr@{}}
\toprule
Policy & Uncompiled & Compiled\\
\midrule
GenPA & $42.73 \pm 0.29$ & $36.36 \pm 0.14$\\
GenPA (flow) & $78.43 \pm 0.44$ & $47.96 \pm 0.56$\\
OpenVLA-OFT & $107.98 \pm 0.85$ & $89.06 \pm 1.08$\\
$\pi_{0.5}$ & $249.58 \pm 0.88$ & $62.02 \pm 0.76$\\
Fast-WAM & $462.32 \pm 8.23$ & $96.56 \pm 0.67$\\
ImageWAM & $226.89 \pm 1.25$ & $96.86 \pm 0.24$\\
\bottomrule
\end{tabular*}
\end{table}
\FloatBarrier

\Needspace{9\baselineskip}
\subsection{{Latency and Memory with Instruction Encoding}}
\label{app:timing-qwen}

{When an instruction requires encoding, GenPA's uncompiled
complete-call latency is $54.90 \pm 0.47$\,ms on H100 and
$75.62 \pm 0.48$\,ms on L40S. These measurements use the same batch-one
LIBERO inputs with Qwen3-4B already loaded and warmed, and include prompt
preparation, tokenization, and text encoding in addition to the policy
computation timed above.}

{For batch-one compiled inference on H100, peak PyTorch allocated
memory is 7.63\,GiB with cached instructions and Qwen unloaded, and
15.18\,GiB with Qwen resident when encoding a new instruction.
Action-only and depth-co-trained policies have the same deployment
architecture and measured memory footprint. These peaks cover post-warm-up
policy calls, excluding loading, compilation, and driver allocations.}

\Needspace{16\baselineskip}
\subsection{{Closed-Loop Performance after Compilation}}
\label{app:compiled-control}

{We re-evaluate {the original training-seed42 regression and flow
policies} on LIBERO and all 10,030 LIBERO-Plus
Clean-Instruction instances after compilation, using the same checkpoints,
the evaluation protocols in Appendices~\ref{app:libero-protocol}
and~\ref{app:plus-protocol}, and the compilation settings used for latency
measurement.
Table~\ref{tab:compiled-control} compares their success rates before and
after compilation.}

\begin{table}[!htbp]
\centering
\caption{{GenPA closed-loop success before and after compilation (\%).
LIBERO-Plus uses Clean-Instruction.}}
\label{tab:compiled-control}
\footnotesize
\setlength{\tabcolsep}{6pt}
\renewcommand{\arraystretch}{1.08}
\begingroup
\begin{tabular*}{\linewidth}{@{\extracolsep{\fill}}llrr@{}}
\toprule
Action adapter & Execution & LIBERO & LIBERO-Plus\\
\midrule
Regression & Uncompiled & 99.10 & 83.79\\
 & Compiled & 99.00 & 83.58\\
\addlinespace[3pt]
Flow & Uncompiled & 98.40 & 82.52\\
 & Compiled & 98.20 & 82.19\\
\bottomrule
\end{tabular*}
\endgroup
\end{table}

\FloatBarrier
\section{Real-world Experimental Details}
\label{app:real-world-details}

This appendix supplements the real-world comparison in
Section~\ref{sec:real-world} with the robot interface, training and
deployment protocol, and evaluation conditions.

\subsection{Robot, Data, and Policy Deployment}
\label{app:real-world-implementation}

\begin{figure}[!htbp]
\centering
\includegraphics[width=\linewidth]{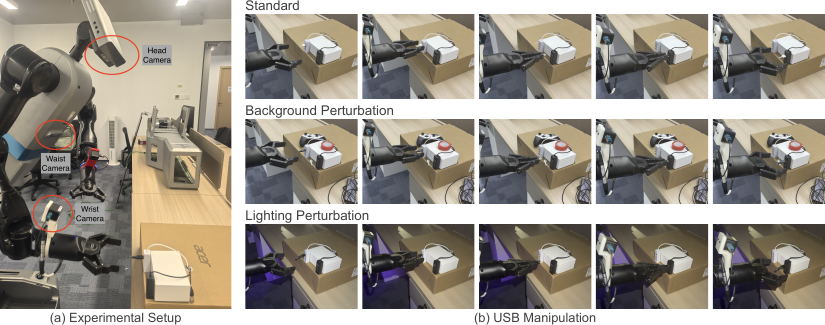}
\caption{\textbf{Real-world setup and USB manipulation.}
(a) The robot platform and camera locations.
(b) Example execution sequences under standard conditions, background
perturbation, and lighting perturbation (top to bottom), progressing from
left to right. The task is to unplug a USB connector from the left hub
and insert it into the right hub.}
\label{fig:usb-real-world}
\end{figure}

\paragraph{{Robot and observations.}}
The experiments use the dual-arm robot shown in
Figure~\ref{fig:usb-real-world}. Only the right arm and gripper are
controlled; the left arm, waist, and head remain fixed. {All policies receive}
{the same} three RGB views from the head, waist, and right wrist,
together with {a language instruction and} a
15-dimensional state comprising seven right-arm joint positions, seven
joint efforts, and the gripper state.
{No policy receives depth at deployment.}

\paragraph{{Demonstrations and training.}}
{We collect 600 demonstrations through teleoperation, with
200 demonstrations for each of three source ports on the left hub.
GenPA, $\pi_{0.5}$, Fast-WAM, and ImageWAM use this shared dataset;
the baselines use their official implementations. GenPA additionally uses
current-depth supervision from depth cameras.} All four
policies are fine-tuned on these demonstrations for 10 epochs and
evaluated at their final checkpoints.
{Fast-WAM and ImageWAM are initialized from
\texttt{Wan2.2-TI2V-5B} and \texttt{FLUX.2-klein-base-4B}, respectively,
with their action branches initialized following the official procedures.
Neither loads a policy checkpoint trained on other robot tasks, and both
retain joint supervision on actions and future RGB observations.
$\pi_{0.5}$ is initialized from the official robot-pretrained
\texttt{pi05\_base} checkpoint.}
{For real-world training, GenPA uses the LIBERO augmentation recipe
specified in Appendix~\ref{app:image-preprocessing}. ImageWAM and
$\pi_{0.5}$ retain their native image augmentations, while Fast-WAM uses
deterministic preprocessing without stochastic image augmentation, as in
its publicly released implementation. The real-world results therefore
compare policies trained with each method's own image augmentation.}

\paragraph{{Policy deployment.}}
{All policies are deployed on the same NVIDIA RTX PRO 6000
Blackwell GPU and use a 50\,Hz controller.
For all policies, arm targets are represented as offsets from the joint
positions observed at the beginning of each chunk and converted to joint
targets for execution. Gripper commands are absolute and binary at
deployment. Table~\ref{tab:real-world-deployment} summarizes the prediction
and execution horizons and flow integration steps. ImageWAM uses its
released Prefix-Only inference configuration, without generating future
images. Each policy executes the number of actions specified in the table,
discards any unused predictions, and replans from fresh observations.
For the baselines, inference and execution do not overlap: the controller
holds the last commanded joint target during inference, without temporal
action averaging or smoothing across chunks.}

\begin{table}[!htbp]
\centering

\caption{Real-world policy deployment settings.}
\label{tab:real-world-deployment}
\small
\setlength{\tabcolsep}{8pt}
\renewcommand{\arraystretch}{1.08}
\begin{tabular}{@{}lccc@{}}
\toprule
Policy & Predicted actions & Executed actions & Flow integration steps\\
\midrule
GenPA & 16 & 16 & ---\\
Fast-WAM & 32 & 24 & 10\\
ImageWAM & 16 & 16 & 10\\
$\pi_{0.5}$ & 50 & 25 & 10\\
\bottomrule
\end{tabular}
\par\vspace{2pt}
{\footnotesize --- denotes direct action regression.\par}
\end{table}

\subsection{Evaluation Protocol}
\label{app:real-world-protocol}

{We evaluate task execution under three visual conditions.
In all conditions, the initial right-arm pose and the source port on the
left hub vary randomly across trials, while the destination remains the
first port on the right hub. At reset, the end-effector position is perturbed
by $\Delta x\in[-0.05,0]$\,m and
$\Delta y,\Delta z\in[-0.05,0.05]$\,m, while its orientation remains fixed.
Each policy is evaluated for 50 trials per
condition, totaling 150 trials.}
\begin{itemize}
  \item {\textbf{Standard:} the nominal scene, with the
  initial-condition variations described above.}
  \item {\textbf{Background:} visual distractors, such as
  a game controller and tape, are added to the workspace.}
  \item {\textbf{Lighting:} room lights are switched off,
  and the scene is illuminated by the recording phone's LED.}
\end{itemize}
{A trial succeeds only if the robot unplugs the connector,
inserts it into the target port, and releases the gripper within 50 seconds
without dropping the connector. {This timeout is used to
terminate stalled trials; in our evaluations, it did not interrupt any
trial still progressing toward task completion.}
We report success rates for each condition
and their arithmetic mean in Table~\ref{tab:usb-draft}.}
Figure~\ref{fig:usb-real-world} shows example executions under each condition.

\FloatBarrier
\section{Generation Details for Figure 1}
\label{app:figure1-generation}

\subsection{Source Image and Illustrative Edits}
\label{app:figure1-source}

\begin{figure}[htbp]
\centering
\begin{minipage}[t]{0.325\linewidth}
\centering
\includegraphics[width=\linewidth]{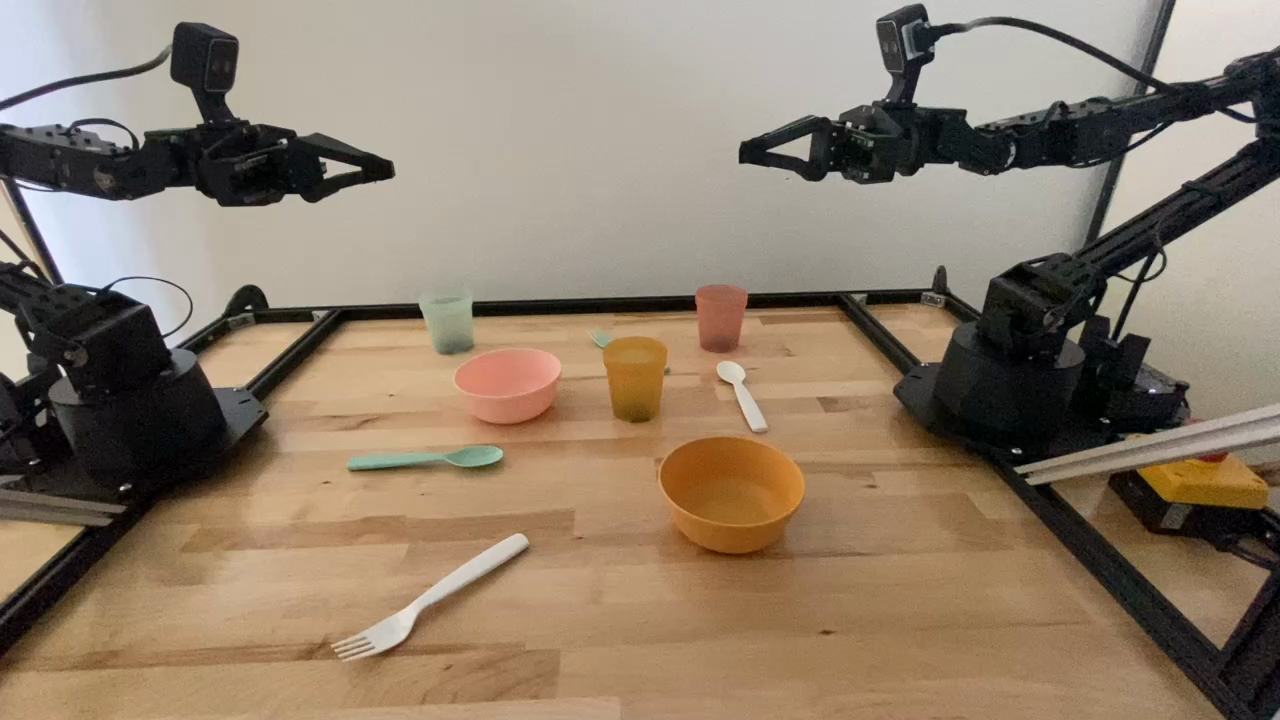}
\par\small (a) Original scene
\end{minipage}\hfill
\begin{minipage}[t]{0.325\linewidth}
\centering
\includegraphics[width=\linewidth]{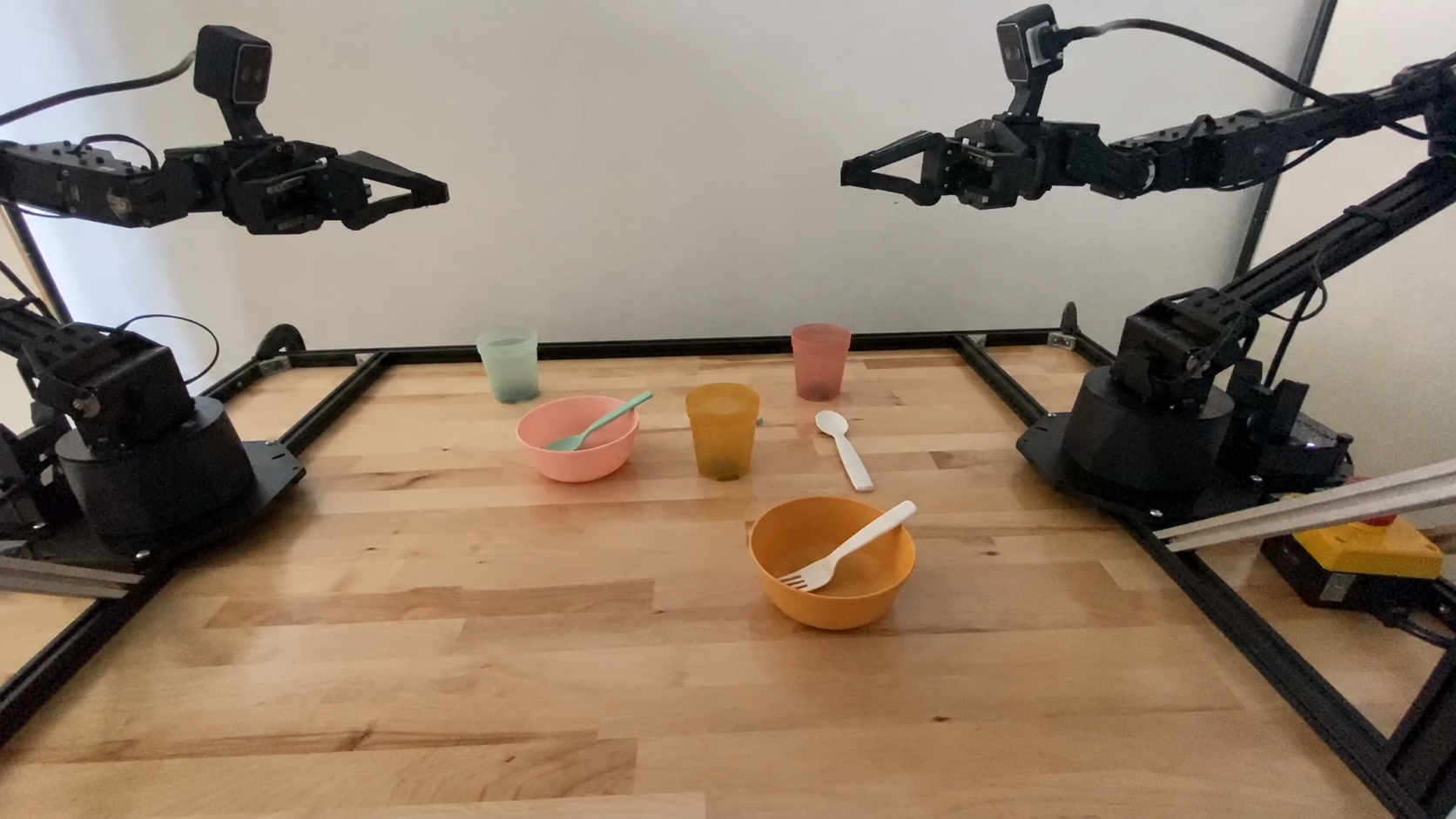}
\par\small (b) Generated goal state
\end{minipage}\hfill
\begin{minipage}[t]{0.325\linewidth}
\centering
\includegraphics[width=\linewidth]{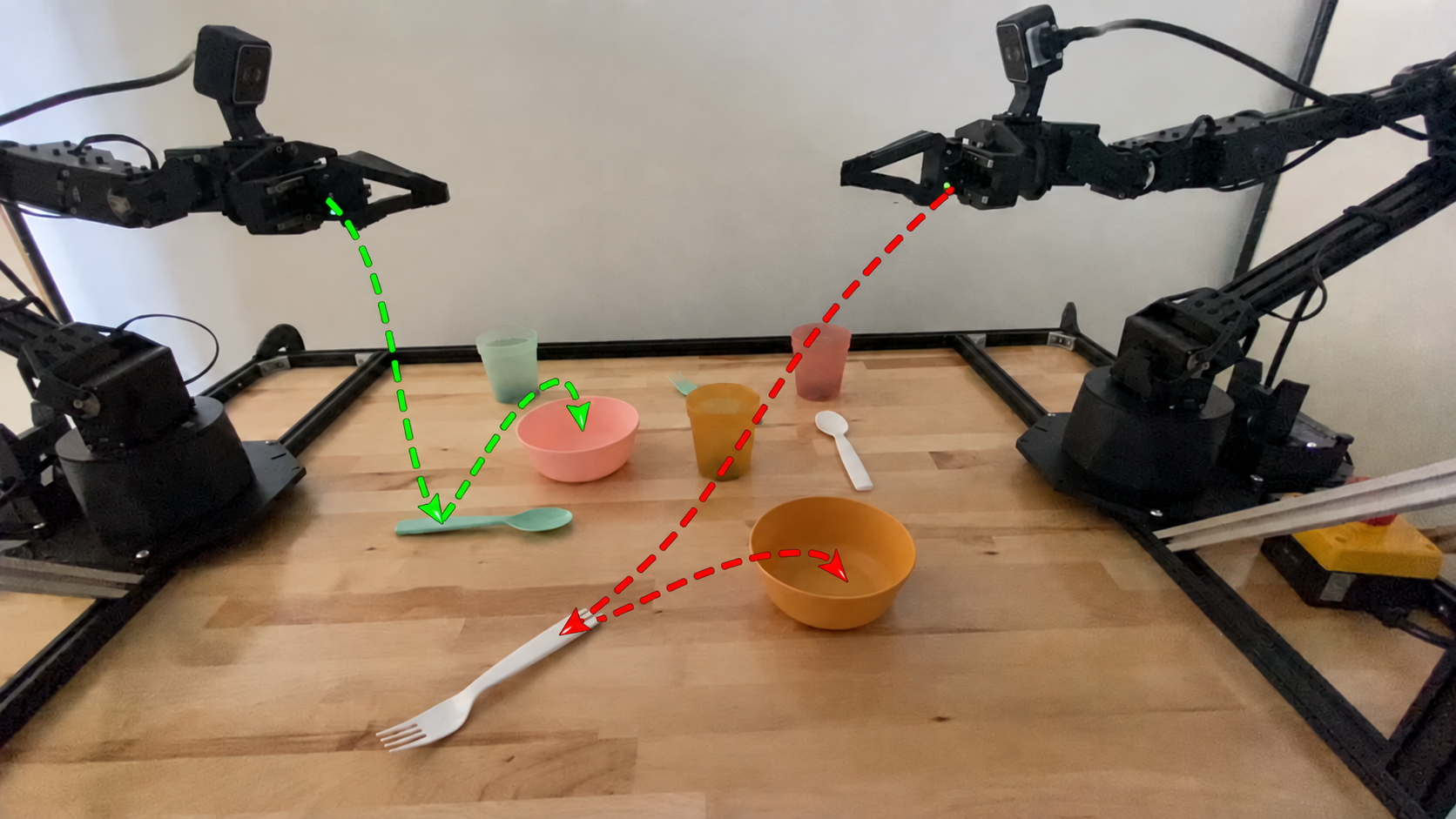}
\par\small (c) Generated motion sketch
\end{minipage}
\caption{{Original scene and generated goal-state and
motion-sketch images used in Figure~\ref{fig:genpa-overview}(a).}}
\label{fig:motivation-generation-details}
\end{figure}

The input is a frame from the \emph{Random Kitchen}
\href{https://aloha-unleashed.github.io/assets/KitchenStack4x.mp4}{source video}
of ALOHA Unleashed~\citep{zhao2025alohaunleashed}.
{Both outputs in Figure~\ref{fig:motivation-generation-details}
were generated from the same source image using GPT Image 2 with separate
prompts. These are illustrative image edits, not GenPA predictions or robot
execution results.} Figure~\ref{fig:genpa-overview}
uses abbreviated prompts for layout; the complete prompts follow.

\FloatBarrier

\Needspace{12\baselineskip}
\subsection{Complete Prompts}
\label{app:figure1-prompts}

\paragraph{Goal-state editing.}
\begin{quote}
Edit this photograph to show the scene after the robot has put the green
spoon into the pink bowl and the white fork into the yellow-orange bowl.
Keep the other utensils, cups, bowls, background, and camera view unchanged.
Each moved object must appear exactly once. Return only the edited scene.
\end{quote}

\Needspace{9\baselineskip}
\paragraph{Image-space motion sketch.}
\begin{quote}
On this photograph, sketch how the robot would pick up the green spoon and
place it into the pink bowl, then pick up the white fork and place it into
the yellow-orange bowl. Use dashed, one-way arrows that follow each
movement from the gripper to the utensil and then to its destination.
Keep the photograph unchanged.
\end{quote}

\end{document}